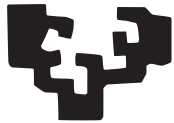

eman ta zabal zazu

Universidad del País Vasco / Euskal Herriko Unibertsitatea

# Euskarazko lehen C1 ebaluatzaile automatikoa

**Egilea:** Ekhi Azurmendi Arrue
**Tutorea:** Oier Lopez de Lacalle Lecuona

## hap

Hizkuntzaren Azterketa eta Prozesamendua Masterreko titulua lortzeko bukaerako proiektua

2024ko urria

---

**Sailak**: Lengoaia eta Sistema Informatikoak

**Laburpena**

Proiektu honetan zehar euskarazko idazlanek C1 maila duten edo ez zehazten duen ebaluatzaile automatiko bat garatzen saiatu gara. Gure helburua betetzeko HABE eta HiTZ arteko hitzarmenaren bitartez 10.000 transkribatutako idazlan eskuratu ditugu gure sistema entrenatzeko. Datu eskasia eta sistemaren gaindoitzea ekiditeko teknika ezberdinak landu ditugu: EDA, SCL eta erregulazioa; Hizkuntza Eredu ezberdinekin ere probak egin ditugu duten portaera aztertzeko. Azkenik, sistema ezberdinen portaeren analisiak ere egin ditugu, ereduen kalibrazioa eta artefaktuen eragina neurtzeko.

# Gaien aurkibidea

# 1 Sarrera

## 1.1 Motibazioa

Gaur egun hizkuntza teknologiak geroz eta erabiliagoak dira, GPT edota Alexa edo Siri bezalako adimen artifizialak oso zabalduta daude gizartean. Teknologia hauek baliabide ugariko hizkuntzetan errendimendu altua izan ohi dute, baina arazoak izan ohi dituzte euskara bezalako hizkuntza txikiagoekin. Euskarak mundu digitalean presentzia edukitzea garrantzitsua da hizkuntza teknologiak euskarara ekartzeko.

Euskaraz teknologia hauek garatzeko eta beharrezko datuak biltzeko IKER-GAITU proiektua sortu zen 2023an. Proiektuak 3 urtez 4 helburu nagusi ditu:

- Idatzizko eta ahozko euskarazko gaitasunak automatikoki sailkatzeko sistema bat garatzea.

- Hizketarako arazoak dituzten pertsonei, ahots pertsonalizatuzko ahots sintesirako teknologien garapena.

- Berehalako ahotsezko transkripzioa, bai euskara eta gaztelaniera nahasten direnean, baita hizlari bat baina gehiagoko egoeretarako.

- Euskararako testuzko hizketarako sistema baten garapena, gaur egungo teknologia onenen pareko kalitatea duena.

Master amaierako proiektu hau IKER-GAITU proiektuaren barruan kokatzen da. Zehazki, lehen helburuko euskarazko idatzien maila zehazteko sistemaren garapena du lan honek helburu.

Idatzizko euskararen mailaren sailkatzaile bat garatzeko, gero eta aukera gehiago dago. Euskaltegietan, hizkuntza-eskoletan eta oro har hezkuntza alorrean sortzen diren hizkuntza-probetako testuak formatu digitalean eskuratzeko eta datu horiek ikerketarako erabiltzeko aukera handiagotzen joan da azken urteetan. Honek, transkribatutako idazlan hauek erabilita, sailkapen sistemak garatzen hasteko lehen saiakerak egitea ahalbideratzen du.

Hizkuntza mailaren sailkatzaileak garatuta daude eta gero eta hobeak dira datu ugariko hizkuntzetarako. Euskaraz berriz, sistema ezberdinak sortzeko saiakerak egin dira, hainbat teknika erabiliz. Horren adibide HiTZ taldeak garatutako lehen C1 sailkatzaile demoa daukagu. Demo hau hobetzearren testu gehiago eskura izanda sailkatzaile hobeago bat garatzen saiatuko gara.

Proiektu honetan garatuko den sistemaren erabilera praktiko bat euskara ikasten ari direnen komunikagaitasun-maila zehaztea da, ikasleen hizkuntza gaitasunak aztertuz. Honek balio izango du ikasleentzat, irakasleentzat eta erakundeentzat; ikasleek haien maila jakiteko, irakasleek ikerketan edo ebaluazioan laguntzeko, eta erakundeek ebaluazio-neurri osagarri bat izateko.

## 1.2 Proiektuko helburuak

Proiektu honen helburu orokorra C1 mailako sailkatzailea garatzea da. Sailkatzaile honek sarreran idazlan bat jaso eta C1 gaindituko duen edo ez aurresan behar du. Horretarako, euskararako dauden HE eta teknologia aurreratuak ikertuko dira.

Proiektuaren barruan hainbat azpihelburu eta ikerketa galdera definitu dira.

Lehena, eskura dauden HE ezberdinekin esperimentuak egitea da, ataza honetan eredu ezberdinak izan ditzaketen eraginak aztertzeko. Azpihelburu honek ikerketa galdera batzuei erantzutea du helburu, besteak beste, HE eleanitzak elebakarrak baina hobeak diren edo ez horrelako atazetan.

Bigarren mailako helburuekin jarraituz, HE egokienekin egindako entrenamenduak hobetzeko hainbat bide esploratuko dira:

**Erregularizazio** teknikak aztertuko dira izan ditzaketen gaindoitzeak eta alborapenak gainditzeko.

**Supervised Contrastive Learning** teknikak esploratuko dira mugan dauden testuak egokiago sailkatzeko helburuarekin.

**Easy Data Augmentation** algoritmoa erabiliko da datu eskasiari aurre egiteko teknika gisa. Teknika hau mota honetako azterketa linguistiko sakonak egiten diren atazetan egokia den edo ez aztertzeko.

Datu eskasia dugunez eta datuak lortzeko transkripzio prozesu bat martxan dagoenez datu gehiago erabiltzearen eragina aztertu nahi dugu. Horrela, datu gehiago erabilita lor dezakegun onura azter dezakegu.

Azkenik, sortutako ereduen portaeraren analisi bat egingo da, sailkatzaileen asmatze-tasak izateaz gain portaerarekin lotutako informazioa eskuratzeko, ikerketa galdera ezberdinak erantzuteko eta HE eta teknika egokienak identifikatzeko.

## 1.3 Dokumentuaren antolaketa

MALaren memoria hainbat sekziotan banatuta dago, lehena hau sarrera izanik. Jarraian, aurrekariak eta memoria ulertzeko behar diren oinarri teorikoak azaltzen dira. Hirugarren sekzioan ingurune esperimentala deskribatzen da: datuak, erabiliko diren Hizkuntza Eredu eta teknikak. Laugarren sekzioan lortutako emaitzak eta ebaluazioak azaltzen dira. Bosgarren sekzioan entrenatutako ereduen analisiak deskribatuko dira. Seigarren sekzioan emaitzetatik eta analisietatik ateratako ondorioak aipatuko dira eta etorkizunerako geratzen diren lanak zehaztuko dira. Azkenik, zazpigarren sekzioa eranskinen atala da, taulak, testuen adibideak eta abar daude bertan.

# 2 Aurrekariak

## 2.1 HE

Sekzio honetan proiektuaren txostena ulertzeko beharrezkoak diren oinarri teorikoak azalduko dira.

### 2.1.1 Hizkuntza ereduak

Azken urteetan Adimen Artifizialak gorakada izugarria izan du eta hainbat ataza burutzeko erabili dira Ikasketa Automatikoa izenez ezagutzen diren teknikak. Ataza horietako batzuk lengoaia naturalaren prozesamenduak (LNP) barne hartzen ditu. Tresna oso erabilgarriak sortu dira LNPren baitan, itzulpen sistemak, txat botak edota sentimendu analisiak egiteko sistemak besteak beste. Teknologia hauen oinarrian hizkuntza ereduak (HE) izeneko eredu estatistikoetan oinarritutako ereduak daude.

HEak testuen (hitz sekuentzien) gaineko probabilitatea kalkulatzen dute. Probabilitate horrek sarrerako testua gizaki batek idatzia izateko duen probabilitatea adierazten du. Probabilitatearen katearen erregela bidez kalkulatzen da Jurafsky eta Martin (2023). $P(W_1 \dots W_n)$ sarrerako sekuentziaren probabilitatearen formula izanik, probabilitatearen formula honela hedatzen da:

$$P(W_{1:n}) = P(W_1)P(W_2|W_1)P(W_3|W_{1:2})...P(W_n|W_{1:n-1}) = \prod_{k=1}^{n} P(W_k|W_{1:k-1}) \quad (1)$$

HEak testuen probabilitateak kalkulatzeaz gain, testu sortzaile moduan ere erabil daitezke, $P(W_i\|W_1 \dots W_{i-1})$ moduko kalkuluak egiteko gai direnez, probabilitate gehien duen hurrengo hitza zein den asma dezake Jurafsky eta Martin (2023).

$$P(\text{zaude}|\text{kaixo, zer moduz}) \quad (2)$$

Honek HEen erabilera esparrua zabaltzen du eta LNP munduko ataza askotan eredu hauen presentzia bermatzen du.

HEak hainbat modutara inplementatzen dira orokorrean. Klasikoki N-grametan oinarritutako ereduak izan dira, non atzetik dauden N-1 hitz kontuan hartuta N-garren hitzaren probabilitatea kalkulatzen duten. Jatorrian atzetik dauden hitz guztiak kontutan hartzen dira, baina probabilitateen kalkulua asko konplikatzen denez, N-grametan oinarritutako sinplifikazioa erabiltzen da. Horri Markoven hipotesia deitzen zaio. Adibidez, 2-gramekin lan egitean, $P(W_i|W_{i-1})$ motako kalkuluak egingo dira. Jurafsky eta Martin (2023)

$$P(\text{kaixo, zer moduz?}) = P(?|\text{moduz}) * P(\text{moduz}|\text{zer}) * P(\text{zer}|,) * P(,|\text{kaixo}) \quad (3)$$

Beste inplementazio bat sare neuronalak erabiliz egin daiteke. RNN motako sareen kasuan RNN (1990), ikusitako sarrerako datu guztien informazioa mantentzen da eta honek

Markoven sinplifikazio beharra ezabatzen du Mikolov et al. (2010). 1. irudian RNN-HE baten diagrama dago, bertan X balioak hitzak izango lirateke, y hurrengo hitzaren probabilitatea eta $h_n$ irakurritako hitzen informazioa gordetzen duen bektorea. $h_n$ bektoreari esker irakurritako informazioa kodetzen da. Halere, sare neuronalen eta RNNen mugengatik errealitatean oso zaila da hitz sekuentziako hitz guztien informazioa mantentzea eta sarrera tamaina handietan arazoak izaten dituzte informazio galerarekin. Jurafsky eta Martin (2023)

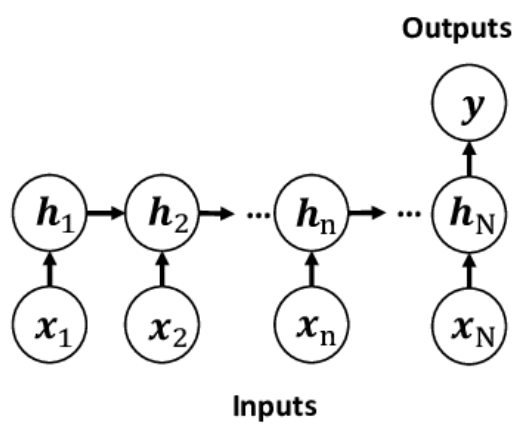

Irudia 1: RNN sare neuronalak HE moduan erabilita.

Sarrerako informazio galera ekiditeko, bi RNN arkitektura berri aurkeztu ziren.

Lehena LSTM (Hochreiter et al 1995) izenekoa 1997an aurkeztu zuten Hochreiter eta Schmidhuber (1997) artikuluan. LSTM arkitekturaren atzean dagoen intuizioa sare neuronal batean modulu gehigarri bat sortzea da, informazioa noiz gogoratu eta noiz ahaztu ikasten duena. Modu horretara RNNek sekuentzia luzeetan duten arazoa txikiagotzen da, informazioa gordetzen duen bektoreak modu selektiboan gordetzen duelako. Gainera, gradienteekin RNNetan zegoen arazoa ere gainditu da, sekuentzia luzeetan atzeranzko barreiaketan gradienteak desagertzen baitziren. 2. irudian LSTM baten diagrama ikus daiteke.

Bigarrena GRU (Chung et al 2014) izenekoa 2014an aurkeztu zuten Chung et al. (2014) artikuluan. Arkitektura honek LSTMaren ideia jarraitzen du, baina sinpleagoa da. Ezberdintasunak daude kalkuluak egiteko moduan bai eta diseinuan ere 3. diagraman ikus daitekeenez. GRUk ere historia kodetzen duen bektoretik zer gorde eta zer erauzi erabakitzen ikasten du LSTMak bezala, baina modu sinpleago batean, parametro eta kalkulu gutxiago eginez.

Nahiz eta GRUk parametro eta kalkulu gutxiago egin, ez du LSTMa ordezkatu eta atazaren arabera arkitektura bat edo bestea aukeratzen da.

LSTM eta GRU erabilita RNNek zeukaten arazoa txikitzea eta HE hobeak sortzea lortu zen. Baina sekuentzia luzeetako arazoak ez ziren guztiz desagertu. Arazoa da sekuentzia osoa bektore batean kodetzen dela eta sarrera luzeak badira, oso zaila da informazio guztia ongi kodetzea. Konponbide gisa Bahdanau et al. (2014) artikuluan arreta mekanismoa

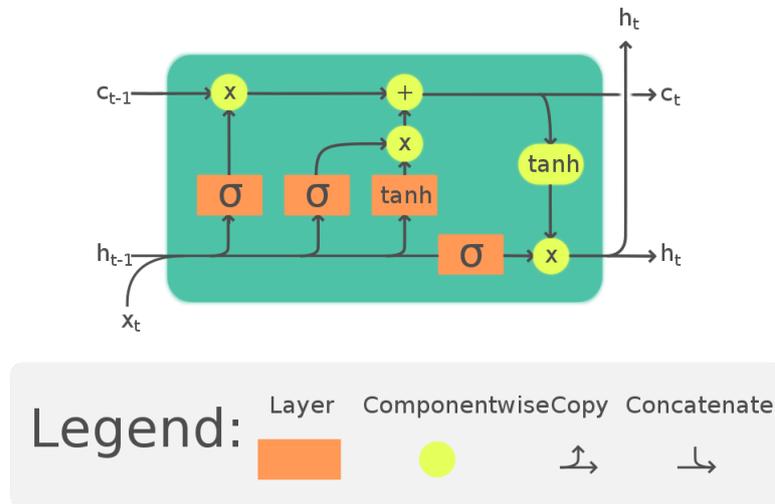

Irudia 2: LSTM baten diseinu irudia.

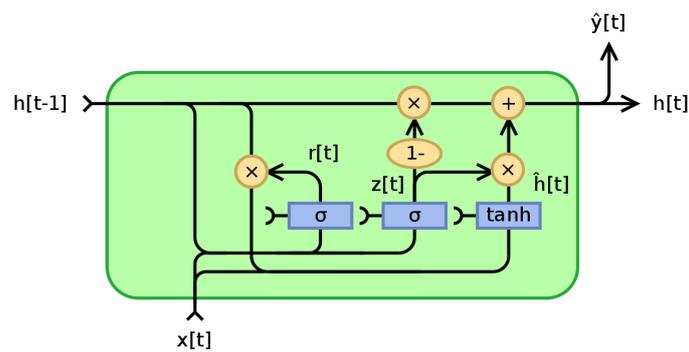

Irudia 3: GRU baten diseinu irudia.

aurkeztu zuten. Honakoa esaten da artikuluan arretaren beharra ulertzeko:

> *A potential issue with this encoder–decoder approach is that a neural network needs to be able to compress all the necessary information of a source sentence into a fixed-length vector. This may make it difficult for the neural network to cope with long sentences, especially those that are longer than the sentences in the training corpus.*

Arreta mekanismoak sarrerako hitz bakoitzaren kodeketa bektoreari eragiketa batzuen bitartez, garrantzia handiagoa edo txikiagoa ematen dio; ondoren, garrantziaren arabera kalkuluak egiteko. Sareak entrenamenduan zehar arreta ze sarrera motan jarri ikasiko du. 4. irudian ikus daiteke arreta mekanismoaren inplementazio diseinua. Ikusten den bezala, sarrerako hitz bakoitzarentzat arreta balio bana kalkulatzen da eta ondoren kodeketa bektoreekin nahastuz irteerak kalkulatzen dira.

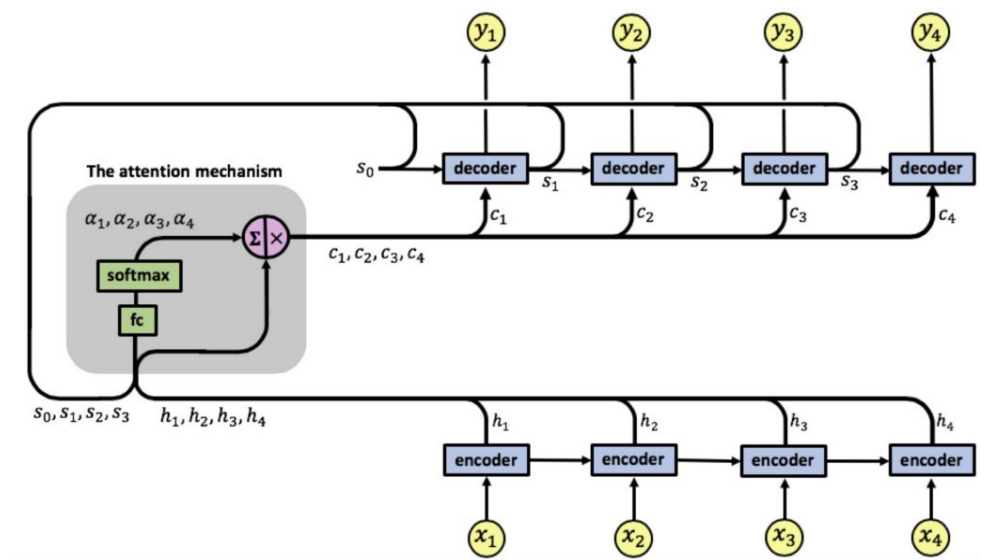

Irudia 4: Atentzio mekanismoaren diseinu irudia.

### 2.1.2 Transformers

2017an *Attention is all you need* artikuluan *Transformers* arkitektura aurkeztu zuten Vaswani et al. (2017). Arkitektura hau arretan oinarritzen da eta sarrerako balio bakoitzari garrantzi bat esleitzen dio ondorengo kalkuluetarako. LNPra ekarriz Izendun Entitateen Identifikazioa (IEI) motatako atazetan lokailuek eta juntagailuek ez dute garrantzirik eta ongi entrenatutako eredu batek arreta bitartez pisu gutxiago esleituko die hitz mota hauei.

Arkitekturaren diseinua 5. irudian ikus daiteke. Sare neuronal hauek bi atal dituzte, bata kodetzaile izenez ezagutzen dena eta sarrerako informazioa kodetzeaz arduratzen dena. Besteak deskodetzaile izena hartzen du eta sarrerako informazioa eta kodetzailetik jasotakoa erabiliz, irteerako probabilitateak kalkulatzen ditu. Kodetzaile eta deskodetzaileak arreta mekanismoa ezberdin kalkulatzen dute Vaswani et al. (2017).

Kodetzaileak autoatentzioa erabiltzen du, honek sarrerako sekuentziako kide guztiak elkarren artean konparatzen ditu eta irteerak eraldatzen ditu atentzio balioak erabiliz. Beraz, sarreran 512 hitz badaude, hitz bakoitzeko 512 arreta balio egongo dira, sarrera bakoitza sarrera guztiarekin alderatzen duelako, bere burua barne. Azkenik, arreta balio guztiak erabiliz, sarrerako hitz bakoitzeko irteerako balio bakarra lortzen da.

Deskodetzaileen kasuan arreta gurutzatua erabiltzen da. Autoatentzioaren antzeko funtzionamendua du, baina $W_i$ hitzaren balioak kalkulatzeko $W_{1:i-1}$ hitzak erabiliko dira. Deskodetzaileek beraz ez dituzte hainbeste atentzio balio kalkulatzen, aurreko hitzen atentzioak bakarrik erabiltzen dituzte eta.

LNP munduan sarrerako datuak tokenak izaten dira. Tokenak HE batek hizkuntza prozesatzeko eta sortzeko erabiltzen dituen oinarrizko testu edo kode unitateak dira. HE gehienek tokenekin funtzionatzen dute, baina karakterekazko HEak ere eskuragarri daude. Testuak tokenetan banantzeko tokenizazio izeneko prozesu bat erabiltzen da. Tokenizazioa gauzatzeko ez dago algoritmo orokorrik eta tokenizatzaile bakoitzak daukan hiztegia erabiliz gauzatzen du tokenizazioa.

5. irudian sarrerako datuak Input *Embedding* izeneko geruza batetik igarotzen dira, bertan, token bakoitza *embedding* batean bihurtzeko. *Embedding-ak* hainbat dimentsiotako bektoreak izaten dira, zati hau entrenamenduan zehar fintzen joaten da eta erlazionatutako hitzak espazio bektorial horretan gertuago kokatzen ditu Liu et al. (2020). 6. irudian hainbat dimentsiotatik 3 dimentsiotara igarotako *embedding* multzoak irudikatu dira.

*Embedding*-ei kodetzaile eta deskodetzaile ataletara sartu aurretik, posizio-kodeketa izeneko balio batzuekin eragiketak egiten zaizkie. Posizio-kodeketa balioek tokenaren posizioari buruzko informazioa ematen die *embedding*-ei; hori egin ezean, hitzen ordenaren informazioa galduko litzateke. Adibidez, autoatentzioa kalkulatzerako garaian, hitzen ordenaren informazioa ez da mantentzen, baina posizio-kodeketa aplikatuta ordena ere kontuan hartzen da. Ghojogh eta Ghodsi (2020)

Jarraian kodetzaile eta deskodetzaileak sakonago azalduko:

**Kodetzailea:** Lehen esan bezala, zati honek sarrerako tokenen informazioa kodetzen du. Irteera gisa ez du irteerako tokenen probabilitaterik itzultzen, informazio kodetua baizik Ghojogh eta Ghodsi (2020). Kodetzaileak bakarrik diren *transformer*-ak testu sailkapenetan edo erregresioetan erabiltzen dira, kodetutako informazioa beste sare neuronal mota batzuetatik igarota. Kodetzaile arkitektura ezagunenetako bat BERT modeloak dira, hainbat atazatan erabiliak izan dira. Adibidez, Liu (2019) artikuluan BERT modelo bat entrenatu zuten estrakzio-laburpenak egiteko.

**Deskodetzailea:** Kodetzaileak ez bezala, deskodetzaile zatiak irteeran hurrengo tokenaren probabilitateak kalkulatzen ditu. Hau da, $P(W_i|W_0....W_{i-1})$ motako kalkuluak itzultzen ditu, hiztegian egon daitezkeen token bakoitzarentzat probabilitate bat esleituz. Ondoren, probabilitate horiek erabiliz hurrengo tokena kalkulatzen da. Testu sorkuntzan oso erabiliak izan dira, baina deskodetzaile zatia bakarrik

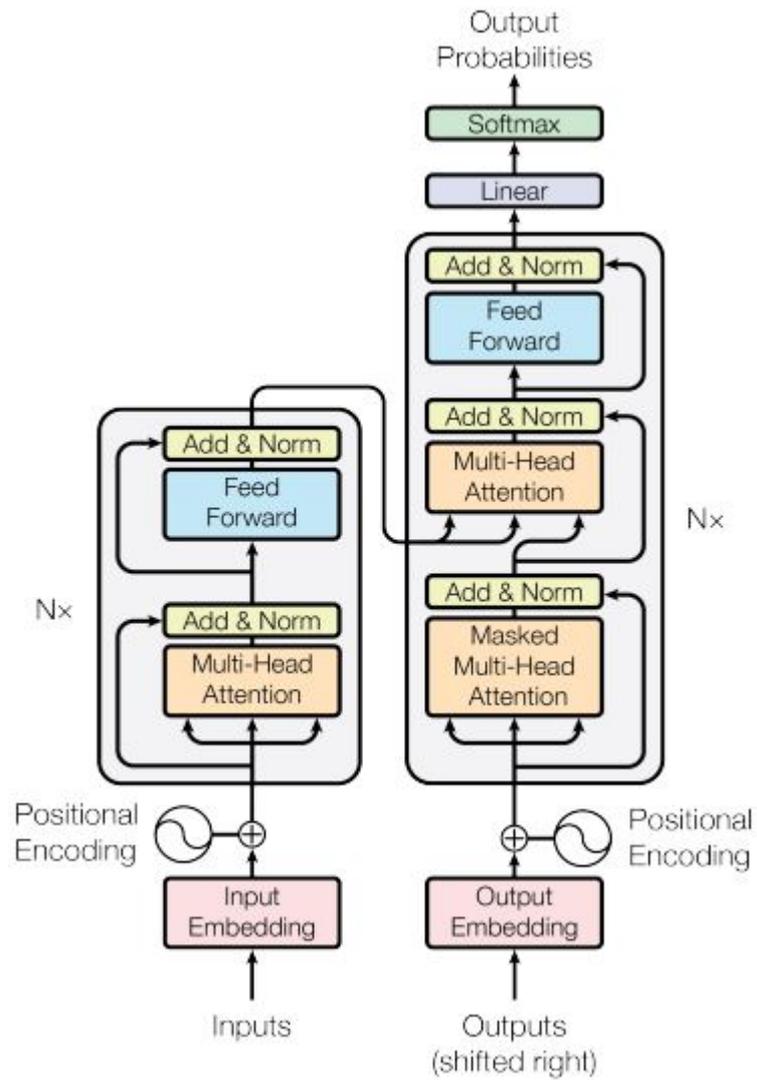

Irudia 5: *Transformers* arkitekturaren diseinu diagrama *Attention is all you need* artikulutik aterata.

Irudia 6: *Embedding* multzoak kasu honetan "airports" hitza aukeratu da eta koloreztatuta agertzen dira entrenamenduan zehar gertuago kokatu diren hitzak. Embedding Projector-etik aterata.

duten *transformer*-ek askatasun handiagoa dute testua sortzerako orduan Ghojogh eta Ghodsi (2020). Azken urteetan oso ezagunak egin diren GPT modeloek Radford et al. (2018), deskodetzaile bat erabiltzen dute azpian.

*Transformer* arkitekturak oso ezagunak eta erabiliak dira LNP atazak burutzeko, baita ere HEak inplementatzeko. Arkitektura honek atentzio balioei esker sarrerako token guztiak kontutan hartu ditzake, probabilitatearen katearen erregela inplementatuz. Hori bai, sarrerako hitz kopurua mugatuta dago RNNetan ez bezala. Baina, RNNekin baina emaitza hobeak lortu dira. Jurafsky eta Martin (2023)

## 2.2 HABEren ebaluazio irizpideak

Euskarak Europako Erreferentzi Marko Bateratua (EEMB) jarraitzen du hizkuntzaren gaitasun mailak sailkatzeko.

Guztira 6 maila erabiltzen dira, maila bakoitza honela definitzen du HABE erakundeak (HABE, 2015):

**A1** Gai da eguneroko harremanetan eta aurrez aurreko testuinguru ezagunetan beharrezkoak diren hitzak eta esapide sinpleak ulertzeko eta erabiltzeko, baita berehalako premiei lotutako galderei erantzuteko ere (bere buruaren eta besteren aurkezpena, eta ezagutzen duen jendeari, haien helbideei eta gauzei buruzko oinarrizko informazioa) .

**A2** Gai da egoera ezagunetan kideekin (lagun, senide, lankide, etab.) aurrez aurreko elkarreraginean gauzen eta pertsonen deskribapenak eta azalpen errazak egiteko eta ulertzeko, baldin eta bere esperientzia-esparruarekin loturiko gaiei buruzko informazio-truke erraza eta zuzena eskatzen badute: bere buruari eta familiari buruzko oinarrizko informazioa, eta erosketei, intereseko lekuei, lanbideei eta abarri buruzkoa.

**B1** Gai da eguneroko testuinguru ezagunetan eta maila bereko solaskideekin elkarrizketak izateko, gai orokorra eta ohikoa izan eta xedea informazioa trukatzea denean. Era berean, gai da lagunei, senideei eta ikaskideei edo lankideei instrukzioak emateko, pertsonak eta objektuak labur deskribatzeko eta gertaeren kontaketa laburrak egiteko. Gai da, baita ere, bere iritziak labur-labur emateko eta bere asmoak azaltzeko.

**B2** Gai da, gai orokor nahiz abstraktuei buruz aurrez aurre dituen solaskideen zein komunikabideetako esatarien testu gehienak ulertzeko, eta ongi bereiziko ditu ideia nagusiak eta bigarren mailakoak. Gai da lagun eta lankideekin, baita jatorrizko hiztunekin ere, ohiko elkarreraginean jariotasunez aritzeko, besteak jakinaren gainean jarriz, iritzia eskatuz, bere ikuspegia defendatuz, etab. Ohiko gai eta egoera ezagunetan adierazpen argiak egingo ditu. Bere lan-esparruko hainbat gairi buruzko deskribapenak eta azalpenak ere emango ditu alderdi esanguratsuak eta xehetasunak bereiziz, eta bere iritzia ere eraginkortasunez emango du.

**C1** Gai da lagunarteko, lan-esparruko eta komunikabideetako edozein testu (elkarrizketak, eztabaidak, azalpenak) ulertzeko; baita inguruko herri-hizkeran sortutakoak ere. Eta gai da, era berean, lagunarteko elkarreraginean, lan-esparruan eta komunikabideetan ideiak eta iritziak jariotasunez eta eraginkortasunez adierazteko. Ongi egituratutako eta osatutako testu argiak eta zehatzak idazteko gai da, ideia nagusiak eta osagarriak bereiziz, eta, oro har, xedea lortzeko estrategia egokiak aukeratuz. Era berean, bere ikuspuntua luze eta zabal adierazteko gai da, ideia osagarriak eta adibide egokiak emanez.

**C2** Gai da norbere espezialitateko nahiz gai orokorretako testuak ulertzeko, eta egoerari egokituz, zuzen eta zehatz, arrakastaz ekoizteko. Gai da, jariotasun handiz eta zehaztasunez, gauzak adierazteko, baita konplexutasun handiko egoeretan esanahiaren nabardura txikiak bereizteko ere.

HABE erakundeak hainbat azterketa dauzka maila bakoitzeko euskara gaitasuna duela zihurtatzen duten zihurtagiriak lortzeko, C1 azterketa horietako bat izanik.

Zihurtagiriak lortzeko azterketa bakoitzak hainbat irizpide linguistiko ditu idazlanak ebaluatzeko. HABEk C1 mailako idazlanak ebaluatzeko honako irizpideak erabiltzen ditu:

**Egokitasuna** Eskatutako gai, solaskide eta xedearen araberako lotura. Solaskideen harreman-motari eta xedeari dagokion tonua. Lexikoaren modalizazioa.

**Koherentzia** Ideia nagusien eta lagungarrien hari logikoa. Diskurtsoaren egitura, garapena eta hurrenkera logikoa. Ideien osotasuna.

**Kohesioa** Esaldi, paragrafo eta testu osoa elkarlotzeko testu-antolatzaileen erabilera eraginkorra. Puntuazio-sistema. Kohesio-mekanismoen erabilera: anaforak, ordezkapenak, elipsiak.

**Aberastasuna** Hizkuntz baliabideen ugaritasuna eta barietatea. Esaldi landuen eta horien baliokideen erabilera. Ordezkapen lexikalen erabilera.

**Zuzentasuna** Gramatika-arauen erabilera zuzena: ortograia, deklinabidea, aditza, morfologia, sintaxia... Gaiari loturiko lexikoa.

## 2.3 Testua ebaluatzeko sailkapen atazak

Testu sailkapenak testu edo hitz-sekuentziari etiketa bat esleitzea du helburu. Formalki jarraiko formularekin deskribatu daiteke testu sailkapena orokorrean.

P funtzio bat izanik, non sarreran testu bat eta etiketa bat emanda, testuak etiketa hori izateko probabilitatea itzultzen duen. $L=L_1, L_2, ...L_n$ atazan definitutako etiketa guztien multzoa izanik eta S sarrerako testua izanik, testu sailkapenak honelako arazoa ebatzi behar du:

$$\arg\max_i P(L_i|S) \qquad (4)$$

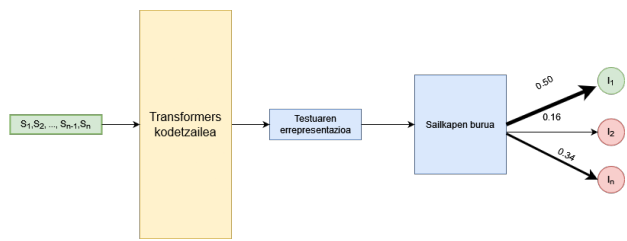

Irudia 7: Transformer kodetzaile bat testu sailkapenerako nola erabiltzen den.

Probabilitate gehien duen etiketa esleitzea da testu sailkapena. P funtzioa oso konplexua da, etiketa kopurua mugatua izanik ere, sarrerako testu ezberdinen kopurua ia mugagabea delako. Hainbat eredu ezberdin erabili dira P funtzioa hurbiltzeko, Naïve Bayes, HMM edota HE neuronalak besteak beste (Li et al., 2022).

Gaur egun, transformer arkitekturaren gorakadarekin, P funtzioa transformer kodetzaileekin hurbiltzen da, nahiz eta deskodetzaileak ere erabiltzen diren. Kodetzaileen kasuan HE gehienek token berezi bat izaten dute sailkapen atazak burutzeko, token horren helburua testuaren informazioa kodetzea da eta azken geruzako errepresentazioa nahikoa izatea dagokion sailkapen atazarako. BERT ereduen kasuan token horrek CLS (Classification hitzaren laburpena) izena hartzen du. Behin CLS tokenak testuak bektore finko batean kodetzen dituela, sailkapen buru bat erabili ohi da etiketa bakoitzari dagokion probabilitatea esleitzeko.

7. irudian sailkapen buru honen funtzionamendua ikus dezakegu. 3 klaseetatik probabilitate gehien lehenari esleitzen dio, beraz, hau izango da sarrerako sekuentziari esleitzen zaion etiketa. Klaseei esleitutako probabilitaten gehiketak 1 balioa eman behar du.

LNPko hainbat ataza testu sailkapen moduan ebazten dira, IEI, testu laburpen estraktiboa eta sentimendu analisia besteak beste. Hala ere, puntuaketak eman behar diren atazetan erregresioa erabili ohi da. Erregresioa sailkapen atazen oso antzeko formulatzen da, baina, kasu honetan klase bakarra egongo da eta ereduek itzultzen duten balioa izango da erantzuna. Probabilitaterik ez denez erabiltzen, irteera balioak ez du zertan [0-1] balio artean egon behar.

HE sortzaileen gorakadarekin sailkapen atazak testuzko erantzunekin ebazteko joera ere zabaldu da. Hau da, kodetzaile eta sailkapen buruak etiketa bakoitzari probabilitate ezberdinak esleitu beharrean, HE sortzaile batek hitzez zuzenean klase bat aurreesatea sarreran agindu egoki bat sartuta. Bestalde, LM Evaluation Harness liburutegia ere oso erabilia da eredu sortzaileak sailkatzaile gisa erabiltzeko. Labur azaltzearren, Harness liburutegian definituta dauden klaseen etiketa hurrengo tokenak izateko probabilitatea begiratzen da eta probabilitate altuena duen etiketaren klasea esleitzen zaio testuari.

HE sortzaileen gaitasunak erabilita entrenamenduko konplexutasunak arindu daitezke, adibide-urriko aginduak emanez entrenatzearen beharra kentzen delako. Baina, deskodetzaileak oso onak izanik ere kodetzaileak emaitza hobeak lortzen dituzte oraindik Etxaniz et al. (2024)-ek erakutsi zuten bezala.

| Enuntziatua | Testu kopurua | Nota | Testuen bb. luzera |
|:-:|:-:|:-:|:-:|
| 1 | 1783 | 2-12 | 350 hitz |
| 2 | 1800 | 1-6 | 350 hitz |
| 3 | 1726 | 0-3 | 150 hitz |
| 4 | 1772 | 0-3 | 150 hitz |
| 5 | 1805 | 0-4 | 150 hitz |
| 6 | 1800 | 0-4 | 150 hitz |
| 7 | 1569 | 0-30 | 250 hitz |
| 8 | 723 | 0-60 | 650 hitz |

Taula 1: ASAP corpusaren deskribapena.

### 2.3.1 Idazlanen puntuaketa automatikoa

Idazlanen puntuaketa automatikoa (IPA), ingelesez Automatic Essay Scoring (AES), oso ataza garrantzitsua eta ikertua izan da urte luzez (Page, 1967, 1968). Definizioz IPA atazak sarrerako testu bati aurrez definitutako puntuaketa sisteman zenbakizko nota bat esleitu behar dio. Ataza hau egoki burutzeko sistema bat garatuta, irakasleen lan-karga arindu daiteke eta ikasleei beraien kabuz ikasteko aukera ematen die, egindako idazlanak automatikoki ebaluatuko direlako.

Hainbat hurbilpen eta datu-multzo ezberdin sortu dira ataza hau burutzeko, izan ere idazlanak ebaluatzeko hainbat irizpide ezberdin erabili dira, narrazio kalitatea edota ideien antolaketa besteak beste (Kumar et al., 2021). Page (1966)-ek lehen hurbilpena hitz kontaketa eta ezaugarri oso sinpleak erabilita egin zuen, gaur egun ordea ikasketa sakoneko teknikak erabiltzen dira ataza honetarako. Taghipour eta Ng (2016)-ek RNN eta sare konboluzionaletan oinarritutako hurbilpenak egin zituzten eta azken saiakerak HEE sortzaileak erabilita egin dira (Stahl et al., 2024).

Corpus ezagunenetako bat Automated Student Assessment Prize (ASAP) da. 2012an aurkeztu zuten Kaggle plataformako lehiaketa batean. 1. taulan corpusaren deskribapena ikus daiteke. Corpus honek 8 enuntziatu ezberdin barne hartzen ditu eta testuak enuntziatuka ezberdinduta daude, gainera, nota ezberdin definitzen da enuntziatu bakoitzeko.

Hainbat hurbilpen diseinatu dira IPA atazak ebazteko. ASAP corpusaren kasuan enuntziatu bakoitzeko eredu ezberdinak sortu izan dira, nota ezberdin ezartzen denez ez baitira bateragarriak. Ohikoena testu sailkapena beharrean, erregresio moduan ebaztea izan da. Adibidez, Tay et al. (2018)-ek hainbat LSTM erabiltzen zituzten arkitektura oso konplexu bat diseinatuz artearen egoerako emaitzak lortzeko.

Azken urteetan transformers kodetzaileak erabili izan dira ataza hau ebazteko. Yang et al. (2020)-ek BERT eredu bat bi motatako galera funtzioekin doitu zuten IPA atazak egokiago ikasteko; alde batetik erregresio bitartez eta MSE galera funtzio erabilita eta beste aldetik ranking moduan ebatzita. Emaitza oso onak lortu zituzten eta garaiko ASAP datu-multzoko artearen egoera bezala finkatu zen.

### 2.3.2 Hizkuntza mailaren sailkatzaileak

IPA atazean testu bati aurrez ezarritako nota sistema batean zenbakizko balio bat esleitzen zaio. Hala ere, IPAn ez da hizkuntzaren maila zehazten. Hau da, testu bat jasota ez da Hizkuntzen Erreferentzia Marko Bateratuan (HEMB) maila zehazten.

Hizkuntza maila zehazteko atazak EEMBko mailekin egiten du lan. Ataza honen hainbat aldaera daude, batzuk testuak maila bat gainditzen duen edo ez zehazten dute, beste batzuk orokorrean A1-C2 tarteko maila bat esleitzen die. Orokorrean ez dago definizio bateraturik ez eta anotazio marko bateraturik. Corpus batzuk maila bat gainditu edo ezgainditu motako anotazioak dituzte (Arrieta et al., 2023), beste batzuk berriz, gaindituko testuen artean maila zehaztuta dute (Geertzen et al., 2013). Ataza hau oso garrantzitsua da hezkuntza munduan, ikasleek beraien idazlanak zuzentzeko sistema automatiko bat izanda, ikasketa prozesuan lagungarria izan daiteke eta irakasleen lankarga gutxiagotzen da. Artikulu ezberdinek garrantzi sistema hauen garrantzia eta idazlanen ebaluazio lan-karga azpimarratzen dute (Zupanc eta Bosnic, 2016).

Datu-multzo ezberdinak sortu dira ataza honekin lotuta. EFCAMDAT (Geertzen et al., 2013) datu-multzo horietako bat da, non ingelesezko L2 ikasleek idatzitako testuak dauden EEMB mailarekin lerrokatuta. Beste datu-multzo bat CLC FCE (Yannakoudakis et al., 2011) izenez ezaguna dena da, bertan 2000 azterketa inguru daude A2tik C1era arte.

Schmalz eta Brutti (2021)-ek BERT ereduak erabilita EEMB maila zehazten zuten EFCAMDAT eta CLC FCE datu-multzoetan oinarrituta. Hurbilpen ezberdinak erabilita ere, ataza hau ez dago konponduta, datu multzo artean ezberdin anotatzen direlako testuak eta ataza beraren definizio zehatzik ere ez dagoelako. Bestalde, ataza oso konplexua da hizkuntzaren ulermen gaitasun sakona behar baita eta datu eskasi eta anotazio marko bateraturik ez dagoenez sistemen garapenean zailtasunak daudelako.

### 2.4 Euskarazko aurrekariak

Hizkuntza mailaren sailkatzaile eta idazlanen puntuaketa automatikoko sistema ezberdinak garatu dira euskararako. Erregela, ezaugarri linguistiko eta ikasketa automatikoko teknika tradizionalak erabilita egin ziren hurbilpen hauek.

Castro-Castro et al. (2008)-ek euskara eta gaztelaniarako IPA ataza burutzeko sistema garatu zuten. Hizkuntzaren ezaugarriak erabiltzen dituzten hainbat modulo definitu zituzten 3 zenbakizko balio itzultzeko. Maila ezberdinetako analisia egin ondoren, 0-10 tarteko 3 balio itzultzen zituen sistemak, jarraiko ezaugarri bakoitzeko bat: ortografia akatsak, lexikoaren aberastasuna eta diskurtsoaren aberastasuna.

Zipitria et al. (2010) eta Zipitria et al. (2011)-ek euskarazko testuak ebaluatzeko sistema garatu zuten erregela eta ezaugarri linguistikoak erabilita. Gehien bat ortografia, gramatika eta lexiko akatsetan oinarritutako ezaugarriak erauzten zituzten. Sistema horrekin adituek maila ezberdinetako eskuz idatzitako laburpenen sailkapena egiten zuten.

Olaizola (2022)-ek IXA ikerketa taldeak HABErekin lehen hitzarmena egin eta lehen hurbilpen baten azterketa aurkeztu zuten. Hurbilpenaren diseinua ezaugarrietan oinarrituta egin zuten, bai ezaugarri tradizionalak erabilita (hitzen karaktere eta silaba kopuruak besteak beste), baita SLA ezaugarri konplexuagoak erabilita (Vajjala eta Meurers, 2012). Artikulu horretan linguagramak erabiltzeko proposamena egin zuten eta A1 mailarako lehen zirriborroa osatu zuten ere.

2023an Arrieta et al. (2023)-ek hitzarmenaren bitartez eskuratutako testuak erabilita sailkatzaile sistema bat garatu zuten. Hurbilpèn hau ezaugarri linguistikoak eta Euskarri Bektoredun Makinak (EBM) erabilita. Hainbat motatako ezaugarriak erabili zituzten: hizkuntza konplexutasun lantzen dutenak, token-, silaba- eta karaktere-kopuruei buruzko ezaugarriak; maila lexikaleko informazio linguistikoa dutenak, aniztasun lexikala eta aberastasun- eta maiztasun-neurriak; maila morfosintaktikoko ezaugarriak, aditz modu-denborak, kasu gramatikalen banaketa edota aditz-formak; ortografia ezaugarriak, akats kopuruak eta motak. 4 sailkapen ataza definitu zituzten, lehena testuak dagokion maila gainditzen duen edo ez, bigarrena, B1etik C2rainoko maila zehaztea, hirugarren testuak B1-B2 edo C1-C2 mailetan dauden zehaztea eta azkenik, HABEren irizpide bakoitzeko nota zehaztea. Emaitza oso interesgarriak lortu zituzten, baina datu eskasia handia dela ere azpimarratzen dute.

# 3 Ingurune esperimentala

Jarraiko atal honetan proiektuan zehar erabilitako datu, eredu, teknikak eta algoritmoak deskribatuko dira ondorengo atalak argiago uler daitezen.

## 3.1 Datuak

Proiektu honetan ikasleen idazlanak ebaluatzeko sistema bat garatu nahi da, zehazki testu bat emanda C1 maila duen edo ez aurresatea da helburu. Ikasketa sakoneko teknika gainbegiratuak erabili ahal izateko, idazlanen lagin bat behar da, C1 maila gainditu duten edo ez etiketarekin. Testu multzo hau HiTZ zentroak eta HABE erakundeak duen hitzarmen baten ondorioz lortu da, beraz, datuak eta idazlanak pribatuak dira eta ezin dira partekatu.

HABEk urtean 2 deialdi egiten ditu C1eko azterketarako. Lehena apirila edo maiatzean izan ohi da eta bigarrena urrian. Deialdi bakoitzean bi idazlanen enuntziatu edo gai aurkezten dira, lehena iritzi testu bat izan ohi da eta bigarrena gutun bat. Deialdi edo epealdiro gaiak ezberdinak izaten dira, beraz, epealdi ezberdinetako idazlanek lexiko edo domeinu ezberdina izango dute, baina idazlanen egiturak mantenduko dira.

Idazlan gehienak eskuz idatzita daudenez, transkribapenak egin behar dira. HABE eta IXAk elkarlanean Hobe_Esk izenez deituko diogun lehen transkribapenak egin zituzten, 614 testu guztira. Ondoren, gainontzeko eskuragarri dauden testu guztiak automatikoki transkribatzeko lana hasi zen eta proiektu honetarako, Hobe_Esk-ez gain 2022ko 9996 testu izan ditugu eskuragarri, 4794 2022ko apirileko deialdikoak (22Api) eta 5202 urrikoak (22Urr).

Automatikoki transkribitu diren ia 10.000 testu horiek, transkribazio akatsak izan ditzakete eta honek sortuko den sistemaren asmatze-tasari eragin oso negatibo bat sor lekioke, zarata egon liteke eta. Zarata hau neurtzeko, transkripzioen 3 errore balio kalkulatu dira:

| Epea  | CER  | Testu konfidantza | Paragrafo konfidantza |
|-------|------|-------------------|-----------------------|
| 22Api | 2.20 | 0.98              | 0.96                  |
| 22Urr | 2.21 | 0.98              | 0.96                  |

Taula 2: Automatikoki transkribatutako testuen errore balioak.

| Testu multzoa | Esaldi kop. | Esaldi luzera | Hitz kop. |
|---------------|-------------|---------------|-----------|
| Hobe_Esk      | 22.47       | 15.73         | 335.87    |
| 22Api         | 23.02       | 15.25         | 333.79    |
| 22Urr         | 22.60       | 15.55         | 335.38    |

Taula 3: Testu multzoen ezaugarriak.

**Character Error Rate (CER)** oker transkribatu diren karaktereen ehunekoa adierazten duena eta testuaren eta paragrafoen konfiantza balioak [0-1] eskalan transkripzioa zein fidagarria den adierazten dute. Epealdika errore balioak 2. taulan ikus ditzakegu. Bai 22Api, baita 22Urr testuetan bataz besteko balioak oso egokiak dira, konfiantza maila oso altua da eta karaktere errore portzentaia oso txikia da, beraz automatikoki transkribitutako testuek ez dute zarata esanguratsurik eta egokiak dira gure sistema garatzeko. Hobe_Esk testuen kasuan eskuz transkribatuak izan direnez, errorik ez dagoela suposatzen da.

Behin sistema garatzeko eskura ditugun datuak izanik, lehen azterketa batzuk egitea komeni da. Egokiena multzo ezberdinetako testuak domeinu ezberdina izanda ere, ezaugarri amankomunak edukitzea da. Ezaugarri hauek 3. taulan konparatzen dira eta orokorrean 3 testu multzoek oso antzeko ezaugarriak dituzte, beraz ez da arazo handirik izango testu ezberdin hauek sistema bera entrenatu edo ebaluatzeko erabilita, oso antzeko ezaugarriak dituztelako.

HABEko zuzentzaileek testuak ebaluatzeko 5 irizpide erabiltzen dituzte, Egokitasuna, Koherentzia, Kohesioa, Aberastasuna eta Zuzentasuna, irizpide bakoitzaren definizioa 2.2. atalean irakur daiteke. Irizpide horiei pisu ezberdinak ezarriz, 0-30 tarteko zenbakizko nota bat kalkulatzen dute eta 15 edo gehiago lortu duten idazlanek gainditu dutela esaten da. Corpuseko 3 multzoek duten notaren distribuzioa, 8. grafikak erakusten digun bezala, 14 inguruan zentratzen da, beraz EZ_GAI klaseko azterketa gehiago ditugu orokorrean. Zehazki Hobe_Esk multzoan batazbesteko nota 14.44 da eta desbiazio estandarra 4.23; 22Api-ren kasuan 13.83 eta 4.53 eta 22Urr-en kasuan 13.31 eta 4.24. 4. taulan epealdi bakoitzeko gainditu eta ez gainditu portzentaiak erakusten dizkigu. EZ_GAI diren testuak gehiengoak direla ikus dezakegu 3 banaketetan, gutxi gorabehera klaseen distribuzioak antzekoak dira, ez gaindituak %60-70 tartean egonik.

Gure corpusean, testuek irizpide bakoitzean lortutako puntuazioak eta azken puntuazioak eskura ditugu, hau interesgarria da sistemen errore analisi sakonagoak egiteko.

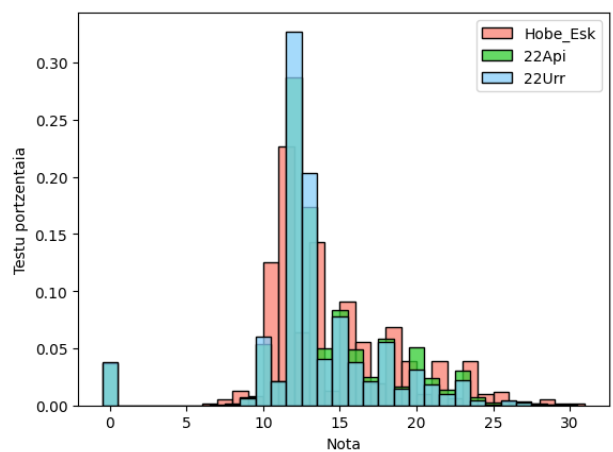

Irudia 8: Testuen kopurua nota bakoitzekiko

| Epea | EZ_GAI % | GAI % |
|---|---|---|
| Hobe_Esk | %60 | %40 |
| 22Api | %63.8 | %36.2 |
| 22Urr | %70.3 | %29.7 |

Taula 4: Azterketa epealdi bakoitzeko etiketen distribuzioa

## 3.2 Hizkuntza Ereduak

C1 sailkatzailea garatzeko euskaraz dakiten 4 HE aukeratu ditugu eta hainbat esperimentu egin dira:

**RoBERTa euscrawl** eredua euscrawl corpusaren gainean entrenatu zen (Artetxe et al., 2022) , Metak garatutako ingeleserako RoBERTa (Liu et al., 2019) ereduan oinarrituta. Transformers arkitekturako kodetzaile zatia den sare neuronal bat da.

**BertEus** eredua BMC corpusaren gainean entrenatu zen (Agerri et al., 2020), Googlek aurkeztutako BERT (Devlin et al., 2018) ereduaren arkitektura erabilita, beraz Transformers kodetzailea da.

**XLM-RoBERTa** (Conneau et al., 2019) RoBERTa ereduan oinarritutako HE bat da 94 hizkuntzetan, euskara barne, aurreentrenatuta dagoena. Hau ere Transformers kodetzailea da.

**Latxa** (Etxaniz et al., 2024) ingelesezko Llama2 (Touvron et al., 2023) eredua euskaraz entrenatuta sortutako HE familia da, 7b, 13b eta 70b parametroko ereduak daude eskuragarri. HE hau Transformers deskodetzaile edo sortzaile bat da.

HE hauek sailkatzaile gisa erabiltzeko, sailkapen buru bat definitzen da ereduaren azken kapako embedding balioak hartu eta MLP sare baten bitartez gure kasuan 2 etiketen

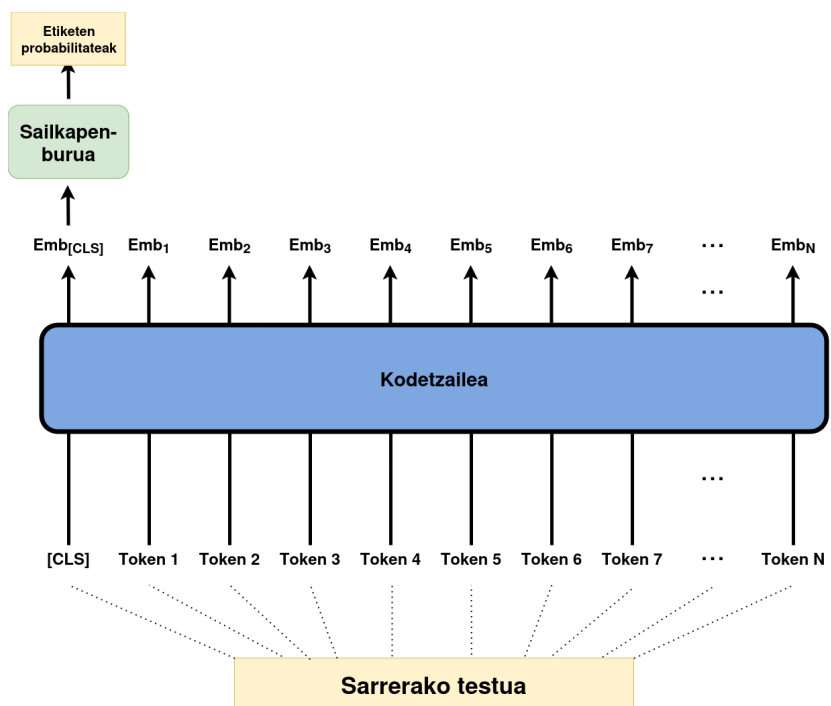

Irudia 9: HE kodetzaile bat sailkapen buruarekin. RoBERTa ereduen kasuan [CLS] tokenaren ordez $<s>$ erabiltzen da.

probabilitateak lortzeko. Kodetzaile diren transformerretan token berezi bat egon ohi da sarrerako testu guztiko informazioa kodetzeaz arduratzen dena, BERT familiako ereduetan [CLS] token berezia dago horretarako eta RoBERTa ereduen kasuan lerro hasiera definitzen duen tokena erabiltzen da $<s>$. Token hauek lehen posizioan egoten dira eta sailkapen buruari token hauei dagozkien azken kapako embedding bektoreak ematen zaizkie sarrera gisa. Deskodetzaileen kasuan azken tokenari dagokion embeddinga da sarrerako testu guztia ikusi duena, beraz sailkatzaile buruari azken tokenari dagokion embeddinga ematen zaio sarrera gisa. 9. eta 10. irudiek kodetzaile eta deskodetzaileak dagozkien sailkatzaile buruekin nola erabiltzen diren irudikatzen dute.

Sailkapen burua ereduaren arabera ezberdin defini daiteke, proiektu honetan erraztasun eta erreproduzigarriak diren esperimentuak egite aldera HuggingFace Transformers (Wolf et al., 2020) liburutegiak eskaintzen duen AutoModelForSequenceClassification objektua erabili da, non sailkatzaile buru hauek aurredefinituta dauden. RoBERTa ereduen kasuan sailkapen burua bi geruzako MLP bat da eta gainontzeko ereduen kasuan geruza bakarrekoa.

### 3.3 Entrenamendu teknikak

Oinarri-lerroen gaitasunak hobetze aldera, teknika ezberdinak probatu dira proiektuan zehar eta jarraian esperimentuak ulertzeko beharrezkoa diren oinarri teorikoak azalduko

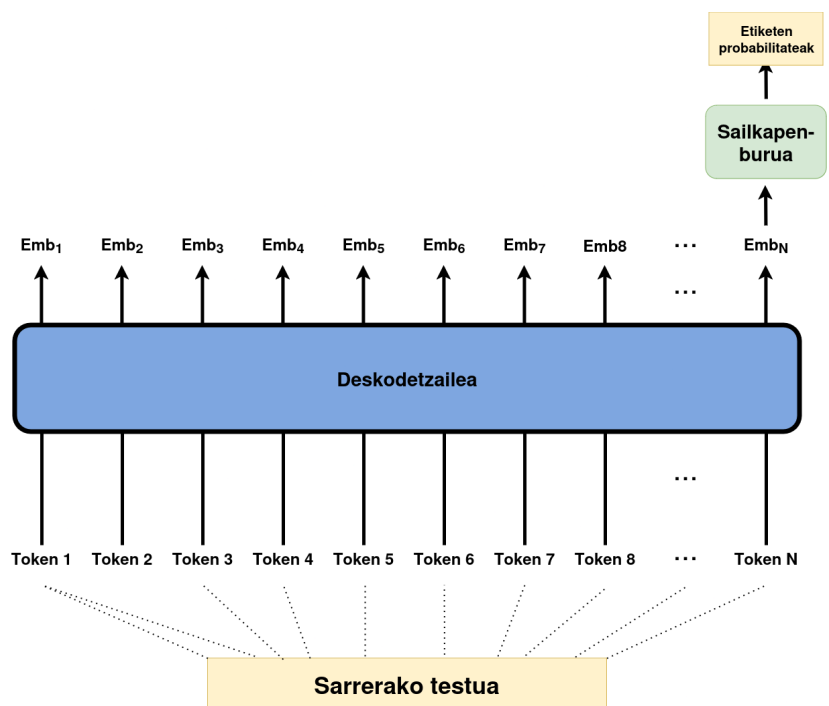

Irudia 10: HE deskodetzaile bat sailkapen buruarekin.

dira.

### 3.3.1 Easy Data Augmentation

**Easy Data Augmentation (EDA)** (Wei eta Zou, 2019) datu eskasiari aurre egiteko lau metodo erraz eta merketan oinarritzen da, jatorrizko testuetatik datu sintetikoak sortzeko. Metodo hauek WordNet erabiltzen dute. EDAk memorizazio eta baliabide gutxiko eszenatokietan entrenatzean sor daitezkeen arazoak saihesten lagun dezake. Jatorrizko artikuluan, egileek teknika hauek erabiliz, baliabide gutxi eta ertainak dituzten datu-multzoetan errendimendua hobetzen dutela erakusten dute. 11. irudian ikusten da EDA erabiltzeak ekar dezaken onura datu multzo txikiak edo ertainak erabilita.

EDAn hurrengo 4 metodoak erabiltzen dira datu gehikuntza egiteko. Algoritmo bakoitzaren eraldaketak ulertzeko jarraiko esaldiarekin sor daitezken eraldaketak erakutsiko ditugu "Gaur mendira joan naiz eta paisaia zoragarria ikusi dut.":

**Sinonimoen ordezkapena (SO):** Eraldaketa honek esaldian hitz hutsak ez diren **k** hitz aukeratzen ditu ausaz. Hitz horietako bakoitza WordNet-eko sinonimo batekin ordezkatzen du. Adibideko esaldian, *paisaia* hitza ausaz aukeratzen bada, adibidez *bista* sinonimoarekin ordezkatuko da, honako esaldi sintetikoa sortuz: "Gaur mendira joan naiz eta bista zoragarria ikusi dut."

**Ausazko Txertaketa (ATx):** Transformazio honek, hitz hutsak ez diren **k** hitz aukeratu eta haien sinonimo bat ausaz txertatzen du esaldiko edozein posiziotan. Adibidera itzuliz, *paisaia* hitza aukeraztuz gero, bere sinonimoa den *bista* esaldiko edozein posiziotan

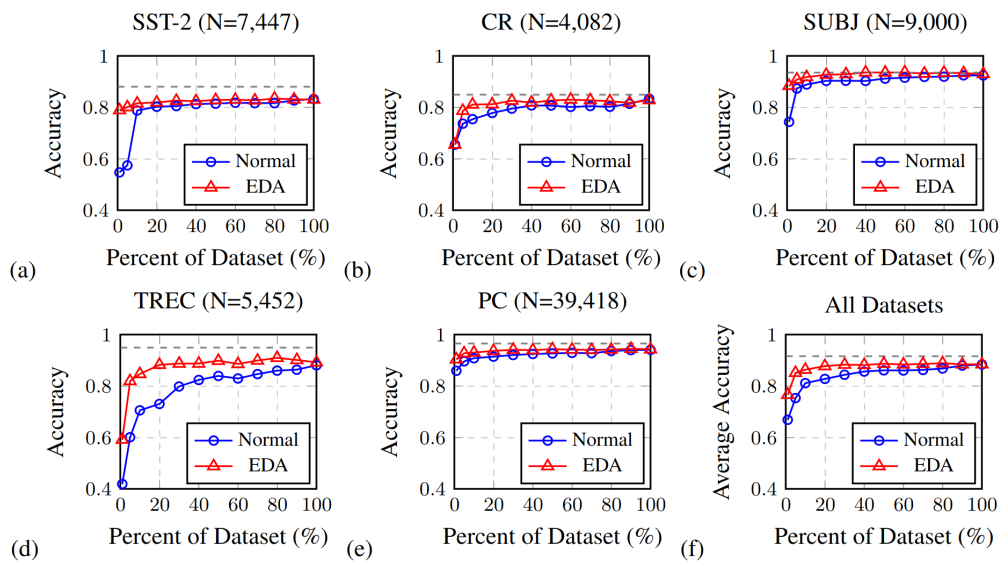

Irudia 11: EDA artikulutik ateratako grafika. Datu multzo txikietan eta ertainetan EDA-ren eragina.

txertatuko du, honako esaldia sortuz: "Gaur mendira bista joan naiz eta paisaia zoragarria ikusi dut."

**Ausazko trukea (AT):** Algoritmo honek esaldian bi hitz ausaz aukeratu eta haien posizioak elkar-trukatzen ditu. Hau **k** aldiz egiten da. Adibidea hartuta, honako esaldia sor lezake: "Gaur paisaia joan naiz eta mendira zoragarria ikusi dut."

**Ausazko ezabaketa (AE):** Esaldiko hitz hutsak ez diren **k** hitz ausaz aukeratu eta ezabatzen dira. Adibideko esaldiari AE aplikatuta, honako esaldi sintetikoa lor genezake: "Gaur joan naiz eta paisaia zoragarria ikusi dut."

EDA teknikek bi hiperparametro dituzte. **n** EDAk sortuko dituen esaldi edo testu berrien kopurua eta $\alpha$ metodo bakoitzaren probabilitatea adierazten du. Algoritmo bakoitzeko k balioak $\alpha$ eta esaldi luzerarekin kalkulatzen dira. Adibidez, $\alpha = 0.1$ erabiliz 10 hitzeko esaldia bada, hitz bakarra eraldatuko da algoritmo bakoitzarekin.

EDAren ingeleseko inplementazioa artikuluaren egileek GitHubeko biltegi honetan utzi zuten atzigarri. Gure kasuan euskarazko testuak erabiliko direnez, euskarara egokitu behar izan dugu. Horretarako NLTK liburutegian eskuragarri dagoen euskararako WordNet bertsioa aukeratu dugu eta euskarazko hitz hutsen zerrenda GitHubeko biltegi honetatik atera dugu. EDA esaldi edo testu labur mailan aplikatu ohi den teknika da, gure kasuan testuak luzeak direnez esalditan banandu dira eta esaldi bakoitzari modu independentean aplikatu zaio gehikuntza. Azkenik, esaldi sintetikoak elkartzen dira testu sintetiko osoa lortzeko.

### 3.3.2 Supervised Contrastive Learning

Constrastive Learning irudiekin lan egitean oso ohikoa den ikasketa teknika da. Teknika honek irudien errepresentazioak espazio bektorialean hurbilago edo urrunago kokatzen

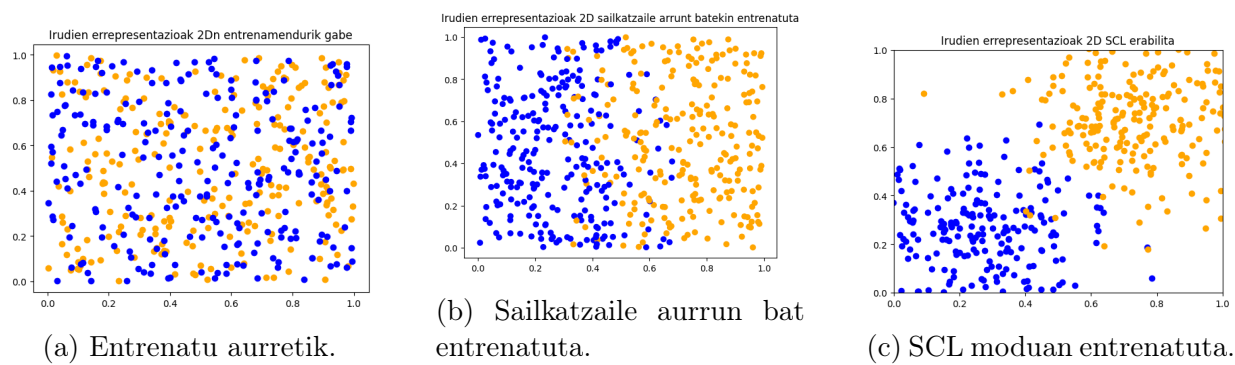

Irudia 12: SCLren eragina modu bisualean.

ditu, atazaren definizio eta helburuaren arabera. Lanketa ugari egin da irudien errepresentazio egokiagoak sortzeko metodo contrastiboak erabilita Chen et al. (2020) eta sailkapen atazetan emaitza oso onak lortu izan dira teknika hauekin.

Supervised Contrastive Learning (SCL) Khosla et al. (2020) hasiera batean irudientzat sortutako ikasketa teknika gisa sortu zuten, artikuluan azaltzen duten bezala. Galera funtzio honen intuizioa irudietatik lortzen diren errepresentazioek klase etiketaren arabera espazioan klusterrak osatzea da. Etiketa bera duten errepresentazioak gertuagotu eta etiketa ezberdinak urrundu egiten ditu funtzio honek. Horrela sailkatzerako orduan espazioan etiketa ezberdinak urrun badaude errazago ezberdinduko dira. 12. irudietan SCLaren intuizioa azaltzeko hiru grafika ikus ditzazkegu: lehen irudian entrenatu aurretik errepresentazioak ausazkoak direla ikus dezakegu; bigarren grafikoan ereduak entrenatu direnez errepresentazioak banatuago daude, baina puntu batzuk ez dira guztiz ongi sailkatu eta banaketa ez da horren argia; hirugarren grafikan berriz multzoak osatzen dituzte, sailkapena errazago eginez.

Irudietaz gain testuekin ere SCLk erakutsi du sailkapen atazetan sistemen errendimendua hobetu dezakeela Cross Entropy Loss (CE) galera funtzioarekin konbinatuz (Gunel et al., 2020). Testuetan ere errepresentazioen gainean lan egiten da, HE batetik ateratako errepresentazioak adibidez, eta hauen kasuan ere multzokatzea da helburu.

5. ekuazioa SCL kalkulatzeko erabiltzen da. $z_i$ balioek, ereduak $x_i$ prozesatu ondoren sortzen duen errepresentazioa adierazten du. $P(i)$ $x_i$-ren etiketa berdinak dituzten instantzien indizeek osatzen duten multzoa da. $A(i)$ berriz, instantzia guztien indizeak, $i$ bera izan ezik, dituen multzoa da. Azkenik, $\tau$ tenperatura adierazten duen eskalar bat da eta $\cdot$ produktu eskalarra da.

Sorta barneko etiketa berdineko instantziak gerturatzen ditu galera funtzio honek eta etiketa ezberdina dituztenak urrundu 13. irudian ikusten den bezala. $\tau$ tenperatura balioa zehazten duen hiperparametro bat da. Tenperatura balio txikiek (zerotik hurbil daudenak) SCLaren eragina areagotzen dute, klase ezberdineko instantzien distantzia handiagotuz eta klase berdinekoen artean txikiagotuz. Tenperatura balio handiekin berriz, SCLren eragina leuntzen da eta distantzia horiek modu leunago batean aplikatzen dira.

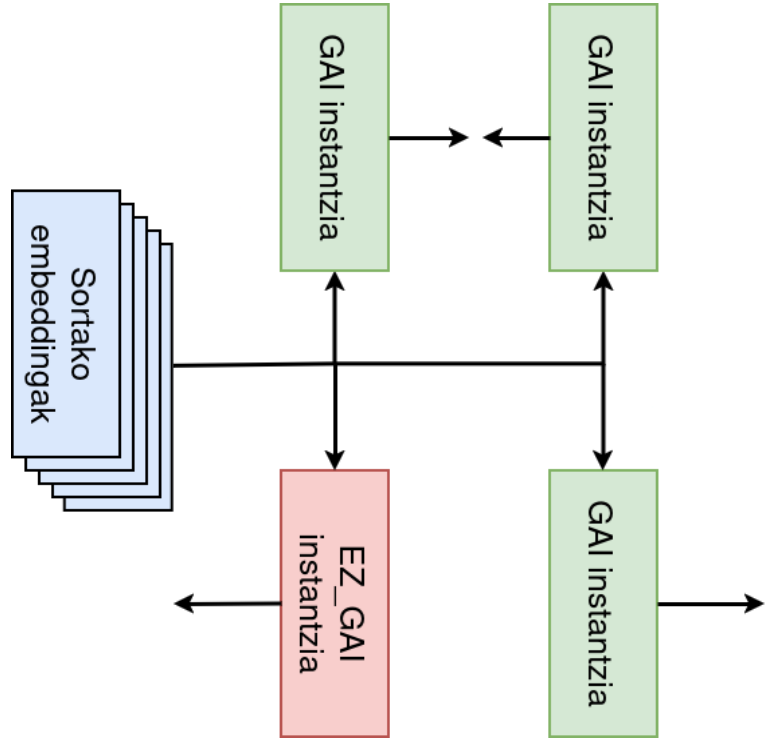

Irudia 13: SCL galera funtzioaren eraginaren azalpena. Etiketa berdineko instantziak gerturatzen ditu eta ezberdinak urrundu.

$$\sum_{i \in I} \frac{-1}{|P(i)|} \sum_{p \in P(i)} \log \frac{\exp\left(\boldsymbol{z}_i \cdot \boldsymbol{z}_p/\tau\right)}{\sum_{a \in A(i)} \exp\left(\boldsymbol{z}_i \cdot \boldsymbol{z}_a/\tau\right)} \qquad (5)$$

SCL eta CE (Cross Entropy) galera funtzioak konbinatu ohi dira, 6. ekuazioan erakusten den bezala, $\lambda$ hiperparametroa galerei pisuak zehazteko erabiliz.

$$(1-\lambda)\mathcal{L}_{CE} + \lambda\mathcal{L}_{SCL} \qquad (6)$$

### 3.3.3 Entrenamenduak azkartzeko metodoak/erabakiak

Proiektuan zehar egin diren entrenamenduek denbora eta kostu konputazional altuak dituzte. Nahiz eta datu kopuru txikia erabili HE txikiekin esperimentu ugari egin dira eta Latxak berriz denbora eta kostu oso altuak ditu exekuzio bakarra egiteko ere. Arazo hauek kontuan hartuta jarraiko optimizazio teknika edota erabakiak hartu dira entrenamenduak azkartu eta kostuak gutxiagotzeko.

**Doitasun gutxiagoko entrenamenduak** Kostuak optimizatzearren doitasun gutxiagoko pisuak erabiltzea egokia izan daiteke memoria eta denbora aurreztearren. Horrelako doitasunak erabiltzea oso egokia da denbora eta memoria aurrezteko.

HAP masterra

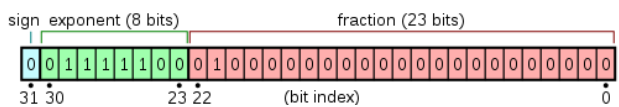

Irudia 14: FP32 doitasunaren definizioa. Wikipediatik ateratako irudia https://en.wikipedia.org/wiki/Single-precision_floating-point_format

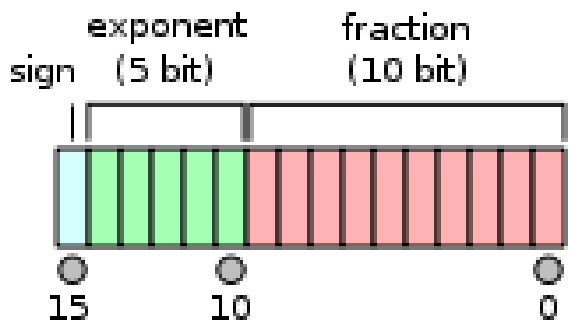

Irudia 15: FP16 doitasunaren definizioa. Wikipediatik ateratako irudia https://en.wikipedia.org/wiki/Half-precision_floating-point_format

Doitasu txikiagoko errepresentazioen gorabeherak ulertzeko lehenik doitasun osoko FP32 ulertu behar dugu. 14. ikusten dugun bezala, FP32 doitasunak 23 bit erabiltzen ditu zenbaki hamartarrak errepresentatzeko, 8 esponentzialerako eta bat zeinurako. FP16-ren kasuan berriz, 15. irudian ikusten den bezala, 10 bit erabiltzen dira hamartarretarako, 5 esponentzialerako eta bakarra zeinurako. FP16 erabilita doitasun gutxiago dugu zenbakiak errepresentatzerako errore gehiago eginez baina memoria eta kostu konputazional gutxiago erabiliz. Hala ere Micikevicius et al. (2017)-ek egindako ikerketan ereduen errendimenduan okertzerik ez duela eragiten eta kasu batzuetan emaitzak hobetzen direla erakutsi zuten, FP16 erabiltzea teknika egokia dela berretsiz.

FP16-k duen aldaera ezagun eta erabili bat BF16 da. FP16-ren oinarri bera jarraitzen du azken honek, baina hamartarretarako 7 bit erabiltzen dira eta 8 esponentzialerako 16. irudian ikus daitekeen bezala. BF16 eta dagozkion teknikak erabilita eragiketak egiteko denbora asko murriz daitekeela ikusi da Osorio et al. (2022). BF16 azken HE erraldoiak entrenatzeko erabiltzen den doitasuna da, adibidez Llama2 eredua Touvron et al. (2023).

Doitasun txikiagoko ereduen onurak ikusita, kodetzaileak FP16 erabilita entrenatuko ditugu eta deskodetzailea BF16.

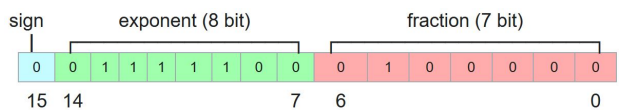

Irudia 16: BF16 doitasunaren definizioa.

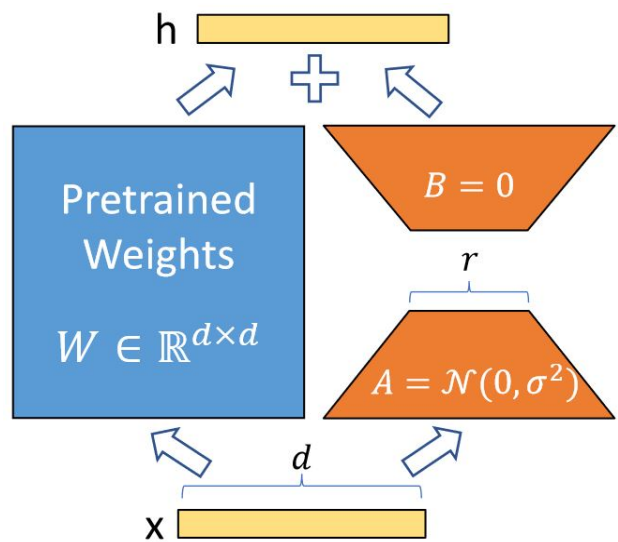

Irudia 17: LoRA metodoaren irudizko azalpena. A eta B matrizeak bakarrik entrenatzen dira.

**LoRA** HE geroz eta handiagoak erabiltzeak errendimenduan onura dakarrela ikusi izan da hainbat ikerketetan. Latxa edota RoBERTa-euscrawl ereduetan emaitzarik onenak parametro gehien dituzten HEek lortzen dituzte (Etxaniz et al., 2024) (Artetxe et al., 2022). Hala ere, parametro kopurua handitzeak bere kostua du, Latxa 70b ereduak 7b-koak baina ia 32 aldiz GPU denbora gehiago behar izan zuen entrenatzeko. Kostu konputazional handi hauei aurre egiteko, Parameter-Efficient Fine-Tuning (PEFT) liburutegia eta LoRA metodoak garatu dira (Hu et al., 2021).

LoRA metodoak ereduaren gainean matrize egokitzaile batzuk sortzen ditu, dimentsionalitate askoz ere txikiagokoak. Oinarrizko ereduen gradienteak izoztu egiten dira eta entrenamenduan matrize egokitzaile hauek egokitzen dira. Inferentzia egiteko oinarrizko eredua eta matrize egokitzailea erabiltzen dira eta bien irteera balioak gehituta lortzen da HE osoaren irteera. Azkenik, doiketak burututa oinarrizko eredua eta matrize egokitzailea elkartu daitezke eredu bakarra osatzeko. 17. irudiak argiago azal dezake LoRA metodoa.

LoRA metodoak hainbat onura ekar dezake entrenamenduan, 17. irudiko A eta B matrizeak bakarrik entrenatzen direnez, entrenatu beharreko parametro kopurua asko jeisten da. Oinarrizko eredua izoztuta mantentzeak memoria erabilera eta konputazio kostua arintzen du, ez baitira pisu horien gradienteak kalkulatu behar. Gainera, metodo honek jatorrizko doitzearen pareko emaitzak edo hobeak lortzen ditu (Hu et al., 2021). LoRA-k hainbat hiperparametro ditu, matrize egokitzaileen dimentsionalitatea eta diluzio balioa besteak beste.

**Sarrera token kopuru muga** Transformer arkitekturako HEek RNNek ez bezala, sarrerako token kopuru maximoa aurretik finkatuta dago eta HEaren sorkuntzan hartzen den erabaki bat da. Kodetzaile gehienetan 512 izan ohi da kopuru maximoa, BERT edota

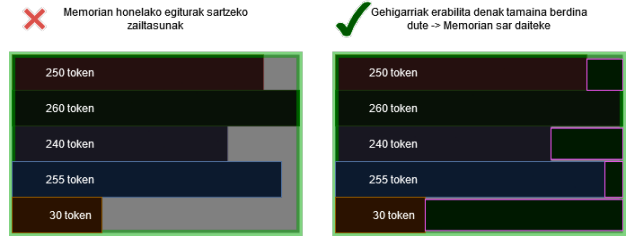

Irudia 18: Sortako instantzien luzerak berdinak izan behar dira.

RoBERTa ereduetan adibidez. Eredu sortzaileetan berriz, handiagoa izan ohi da, Llama2 edota Latxa ereduan 4096koa da Touvron et al. (2023) Etxaniz et al. (2024).

Gure atazean testu oso luzeak ditugu, baina Latxa ereduekin lan egitean sarrerako testuak ez dira 4096 tokenetara iristen, beraz, sarrera mugatzea lagungarria izan daiteke memoria aurrezteko. 18. irudiak erakusten duen bezala, sarrerako token kopurua berdina izan behar da sorta guztian zehar, ezberdinak diren instantziak sorta berean sartu ahal izateko, esanahirik gabeko token betegarri bat erabili ohi da. Hardware eta kodeketa kontuengatik token sarrera 2-ren berredura den luzerakoa izan behar du memoria erabilera aurrezteko.

19. grafikak erakusten digu RoBERTa-euscrawl, BERTeus, XLM-RoBERTa eta Latxa tokenizatzaileak erabilita esaldietako token kopurua. Euskarazko bi kodetzaileetan (BERTeus eta RoBERTa-euscrawl) eta XLM-RoBERTa ereduan 512 inguruko token kopuruan dabiltza testuak, beraz, ezin da sarrera kopurua mugatu. Latxaren kasuan berriz, 4096 du sarrera maximoa eta gure testuak orokorrean 1000 token azpian dabiltzanez dabiltzanez, 1024ko ($2^{10}$-eko) muga ezarri dezakegu.

Azpimarratzekoa da XLM-RoBERTa eta Latxa ereduen tokenizatzaileek token gehiago behar dituztela gure testuak errepresentatzeko, tokenizatzaileak euskara ez den beste hizkuntzak kontuan hartuta egin zirenez, euskarazko hitzak kodetzen zailtasunak dituzte.

## 3.4 Ebaluazio metrika automatikoa

Ataza hau sailkapen ataza bat izanik asmatze-tasa erabiliko dugu metrika automatiko gisa 7. ekuazioan ikusten bezala kalkulatuz. Azpimarratzekoa da sailkapen bitarra denez, F1-micro eta asmatze-tasa baliokideak direla.

$$AT = \frac{EP + EN}{EP + EN + PF + NF} \quad (7)$$

AT = Asmatze-tasa; EP = Egiazko positiboa; EN = Egiazko negatiboa; PF = Positibo faltsua; NF = Negatibo faltsua.

Gure testuak GAI eta EZ_GAI etiketez gain zenbakizko nota bat dute, beraz, nota oso baxuko, nota ertaineko eta nota oneko testuen banaketa egingo dugu sistemaren gaitasunak egokiago aztertzeko gai izateko. Ondorioz, 4 AT balio izango ditugu, bat banaketa

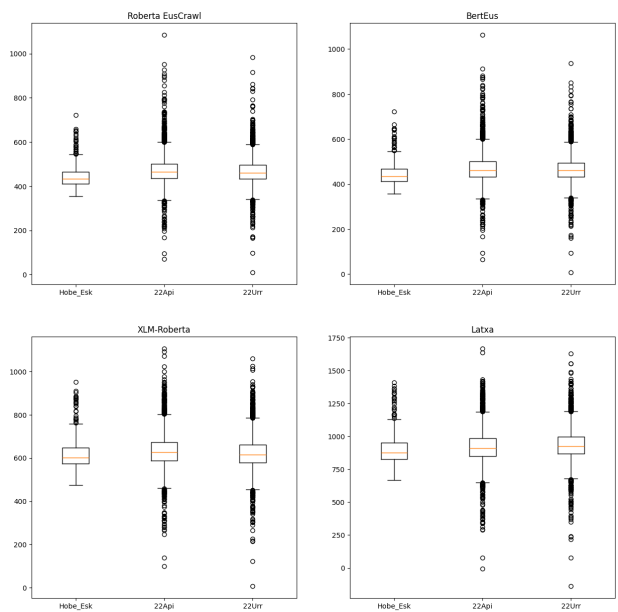

Irudia 19: HEezberdinen tokenizatzaileen token kopuruak bakoitzeko eta azkena orokorra. Gure helburua sistema ahalik eta orekatuena izatea denez, AT orokorraz gain banaketetako ATei ere erreparatu behar diegu sistema egokiena aurkitzeko.

Banaketa hauek egiteko 8. grafikako datuen distribuzioa kontutan hartu dugu eta horrela egin dugu banaketa:

**AT Baxua** Banaketa honetan nota oso txarra, 14 baina gutxiagoko nota, duten testuak sartuko dira. Hau da, $Nota < 14$.

**AT Duda** Banaketa hau mugan dauden testuekin osatuko dugu, beraz 14 eta 16 nota tarteetan dauden testuak egongo dira. $14 \leq Nota \leq 16$.

**AT Altua** Banaketa honetan nota oso onak dituzten testuak egongo dira, hau da, 16 baina gehiago duten azterketak. $16 < Nota$.

# 4  Esperimentuak eta emaitzak

1. atalean deskribatu den sailkatzaile sistema garatzeko, 3. atalean deskribatutako ikasketa sakoneko teknikak eta HEak hartuko ditugu kontuan. Jarraiko orrialdeetan ingurune esperimental honekin egindako esperimentu ezberdinekin lortutako emaitzak eta ateratako ondorioak deskribatuko dira. Esperimentu bakoitzaren banaketa guztietako asmatze-tasak eta hiperparametroak A. ataletan daude ikusgai.

## 4.1 Domeinuaren eragina eta datu-multzoen banaketa

Gure helburua betetzeko eskura dugun datu kopuru txikia kontuan hartuta, lehenik ingelesa bezalako baliabide ugariko hizkuntza batean esperimentuak egin ditugu, bertan ikasitako eta ateratako ondorioak gure domeinu eta hizkuntzara ekartzeko. Ingelesezko probetarako, EFCAMDAT corpusa erabili dugu, bertan ingeleseko L2 ikasleek idatzitako hainbat idazlan daude. Idazlan guztiak gaindituta daude eta A1 mailatik C2ra banatuta daude. Maila bakoitzerako 24 enuntziatu daude eta mailen artean ez dira partekatzen, hau da, A2 idazletarako egin zen galdera edo gaia ez da C1eko azterketan egingo, ondorioz, maila bakoitzeko idazlanak domeinu ezberdinetakoak dira.

EFCAMDAT corpuseko guztira 255316 testu erabili genituen eta ausaz 3 zatitan banatu genituen, entrenamendu, ebaluazio eta test banaketa %70/%15/%15 izanik. Egindako proba guztietan testeko asmatze-tasa %99 ingurukoa izan zen, bai eredu handiak erabilita, ingelesezko RoBERTa large eta XLM-RoBERTa-large, baita eredu oso txikiak erabilita ere, bert-tiny (Turc et al., 2019). Emaitza hauen analisia egin ondoren azterketa gaien domeinuak eragina zuela entrenamenduan ikusi genuen eta hori egiaztatzeko gaiak kontuan hartuta datu banaketa berri bat egin genuen. Oraingoan emaitzak oso txarrak izan ziren, asmatze altuenak %40 ingurukoak izanik eta azterketa gai edo domeinuaren eragina handia zela neurtu genuen. Ingeleseko esperimentu hauetatik, euskarazko gure corpusean domeinuak garrantzia izan lezakeela ondorioztatu genuen eta esperimentuetan kontuan hartu beharko zela.

Domeinuaren eragina kontuan hartuz eta eragina leundu nahian, gure corpusean 2 datu-multzo banaketa egin ditugu:

**Nahastuta** Lehena, entrenamenduan eskura ditugun 2022ko 2 epealdiak agertzen dira. Domeinuaren eragina handia bada, ingeleseko esperimentuetan bezala, emaitza hobeak izan beharko lirateke.

**Epeka** Bigarrena berriz, gaiak kontutan hartutako banaketa, gure kasuan badakigu azterketa epeetako gaiak ez direla errepikatzen, beraz, epealdika banatzen baditugu entrenamendu eta testeko domeinuak ezberdinak izango dira.

Jarraiko 5. taulan erabilitako 2 datu-multzo banaketak deskribatzen ditu. Taulan ikus daitekeen moduan, domeinuaren eragina sakonago aztertzeko, entrenamendu Barne test bat osatu dugu, guztira 794 testuk osatzen dutena. Bi banaketetan ebaluazio, test eta barne-test zatiak konparagarriak dira, testu berberak direlako. Test banaketaren kasuan guztiz domeinutik kanpoko testuak dira, bi banaketetako entrenamenduan ez baitago Hobe_Esk delakoaren testurik. Klaseen distribuzioa berriz, 6. taulan ikus dezakegu eta orokorrean antzekoa da banaketa guztietan, honek distribuzio arazoak ekidingo dizkigu ikasketa prozesuan.

## 4.2 HE aukeraketak eta oinarri lerroak

Ingeleseko esperimentuen ondorioekin osatutako bi datu-multzoak izanik, HE egokienak aukeratzeko esperimentuak egin ditugu. Lehen esperimentu hauen helburua HE ezberdinek

| Banaketa | Entrenamendua | Garapen | Testa | Barne testa |
|---|---|---|---|---|
| Nahastuta | 4000 22Api + 2000 22Urr | 2000 22Urr | 614 Hobe Esk | 794 22Api |
| Epeka | 4000 22Api | 2000 22Urr | 614 Hobe_Esk | 794 22Api |

Taula 5: Esperimentuetarako egindako bi datu-multzoen banaketak.

| Banaketa (EZ_GAI / GAI) | Entrenamendua | Garapen | Test | Barne-test |
|---|---|---|---|---|
| Nahastuta | 0.66 / 0.34 | 0.7 / 0.3 | 0.6 / 0.4 | 0.64 / 0.36 |
| Epeka | 0.64 / 0.36 | 0.7 / 0.3 | 0.6 / 0.4 | 0.64 / 0.36 |

Taula 6: Datu-multzoen klase distribuzioa.

izan dezaketen eragina aztertzea da.

3. atalean deskribatutako 3 kodetzaile eta deskodetzaile bat erabiliko ditugu HE moduan. Eredu hauen BertEus base (125M parametro), XLM-RoBERTa-large (561M parametro) eta RoBERTa-euscrawl large (355M parametro) bertsioak erabiliko dira zehazki, dagozkien sailkapen buruekin. Eredu bakoitza bi datu-multzotan entrenatuko da, HEak konparatzeaz gain datu-multzoen eragina ere neurtzeko. Esperimentu konparagarriak izan daitezen hiperparametro berdinak erabili dira esperimentu guztietan A apendizean ikus daitezkeenak.

7. taulan ikus daitezke HE ezberdinekin lortutako garapen ataleko emaitzak. 3 kodetzailetatik emaitzarik onenak RoBERTa-euscrawl ereduak lortu ditu. XLM-RoBERTa eleanitzak beti EZ_GAI etiketa aurresataten bakarrik ikasi duela dirudi. Agian ataza hau burutzeko ezaugarri linguistiko sakonak kontuan hartu behar direnez eleaniztasun horrek trabak ezartzen dizkio, gainera euskara gutxi ikusi badu entrenamenduan zailtasunak izango ditu testuak sakonki aztertzean. BERTeus ereduak emaitza txukunak lortu baditu ere, RoBERTa-euscrawl hobea da orokorrean. Bi datu-multzoen artean 3 HE ereduetan egokiagoa da Epeka banaketa erabiltzea Nahastutakoa baino, azken honek beti EZ_GAI itzultzeko joera handitzen baitie ereduei.

Kodetzaileekin esperimentuak egin ondoren, Latxa deskodetzailea probatu dugu, konputazio kostu handiak direla eta 7Bko ereduak bakarrik entrenatu ditugu eta kostuak murriztearren, PEFT Lora metodoak erabili. Emaitzak 7. taulan ikus ditzakegu, Latxa ereduak ez dira RoBERTaren emaitzetara gerturatzen. Datu-multzoak konparatuta, egokiena berriro ere Epeka egitea dela ikus dezakegu, emaitza orekatuagoak lortzen direlako. Oro har, ikus dezakegu gehienetan EZ_GAI etiketa itzultzeko joera handiagoa duela Latxak, are eta gehiago Nahastutako datu-multzoan RoBERTa ereduekin gertatzen den bezala. 18. taulako Latxaren entrenamenduko AT balioak aztertuta, joera oso antzekoa ikusten dugu, EZ_GAI diren testuak oso ongi sailkatzeko gai da baina gainontzekoen kasuan txarto egiten du.

4 HEetan entrenamenduko kasuak oso ongi sailkatzen dituzte, baina garapenekoak okerrago, hau da, gaindoitze bat gertatzen da entrenamenduetan eta erregularizazio falta

| Banaketa | Eredua | AT Baxua | AT Duda | AT Altua | AT |
|---|---|---|---|---|---|
| Nahastuta | BERTeus | 0,82 | **0,51** | **0,70** | 0,76 |
| | XLM-RoBERTa | **1,00** | 0,22 | 0,00 | 0,70 |
| | RoBERTa-euscrawl | 0,95 | 0,33 | 0,47 | **0,78** |
| | Latxa-Lora | 0,97 | 0,24 | 0,25 | 0,74 |
| Epeka | BERTeus | 0,90 | 0,45 | 0,51 | 0,77 |
| | XLM-RoBERTa | **1,00** | 0,20 | 0,00 | 0,70 |
| | RoBERTa-euscrawl | 0,85 | **0,50** | **0,65** | **0,77** |
| | Latxa-Lora | 0,95 | 0,29 | 0,27 | 0,74 |
| | Majority Class | 1.0 | 0.14 | 0.0 | 0.70 |

Taula 7: 3 kodetzaileen eta Latxa ereduaren probak bi datu-multzotan. Beltzez eredu orekatuena. Emaitzak garapen atalekoak dira.

nabaritzen da. 4.3.1. atalean arazo hau konpontzen saiatuko gara.

HEen aukeraketa esperimentuetan, RoBERTa-euscrawl eredua egokiena dela ikusi da, BertEus ez oso urrun egonik. XLM-RoBERTa eredu eleanitzak berriz, beti EZ_GAI itzultzen bakarrik ikasi du eta ondorioz eleaniztasunak mesederik egiten ez duela ondorioztatu dugu. Latxaren kasuan berriz, EZ_GAI itzultzeko joera handia dagoela ikusi da eta hau erregulazio faltagatik izan daitekeelako hipotesia dugu, hurrengo esperimentuetan egokiago aztertuko da.

Bi datu-multzoren alderaketak eginda, Epeka egiteak emaitza hobeak lortzen direla dirudi bai Kodetzaile ereduekin baita Deskodetzaileekin ere. Gainera, Nahastutako datu-multzoan entrenatutako ereduek EZ_GAI itzultzeko joera handiagoa erakutsi dute.

## 4.3 Hobetzeko esperimentuak

Oinarri-lerro gisa RoBERTa-euscrawl Epeka datu multzoan entrenatutakoa dugu, emaitzarik onenak eta orekatuenak lortzen dituelako. Eredu hau oinarri gisa hartuta, errendimendua eta orokortze gaitasunak hobetze aldera jarraiko esperimentuak egin ditugu. Erregularizazioko hobekuntzak Latxa ereduarekin ere egin dira, baina, gainontzekoak konputazio kostua dela-eta RoBERTa-euscrawl ereduarekin bakarrik egin dira, eredu honekin hobekuntza nabaririk egonez gero Latxa ereduarekin ere halakoxeak izango direlaren hipotesia izanik.

### 4.3.1 Erregularizazio tekniken esplorazioa

4.2. atalean egindako esperimentuetan, entrenamendu multzoan gaindoitze handi bat gertatzen da. 8. taulan ikusten den bezala, entrenamenduko AT balioaren garapeneko ATaren arteko ezberdintasuna oso handia da RoBERTa-euscrawl ereduaren kasuan, Latxarenean berriz oso tarte txikia da, baina 4.2. atalean azaldu den bezala, ikasketak aurrera

| Banaketa | Eredua | Train AT | Ebal AT | $\delta$ |
|---|---|---|---|---|
| Nahastuta | RoBERTa-euscrawl | 1.0 | 0.78 | 0.28 |
| | LatxaLora | 0.78 | 0.74 | 0.04 |
| Epeka | RoBERTa-euscrawl | 0.93 | 0.77 | 0.16 |
| | LatxaLora | 0.78 | 0.74 | 0.04 |

Taula 8: Entrenamendu eta garapen arteko ezberdintasuna ebaluazioan.

| Banaketa | Eredua | Diluzio-p | AT Baxua | AT Duda | AT Altua | AT |
|---|---|---|---|---|---|---|
| Nahastuta | RoBERTa-euscrawl | - | 0,95 | 0,33 | 0,47 | 0,78 |
| | | 0.3 | 0.9 | 0.47 | 0.57 | 0.79 |
| Epeka | RoBERTa-euscrawl | - | 0,85 | **0,50** | **0,65** | 0,77 |
| | | 0.3 | **0.91** | 0.37 | 0.58 | **0.79** |
| Nahastuta | LatxaLora | - | **0,97** | 0,24 | 0,25 | 0,74 |
| | | 0.3 | 0.87 | **0.43** | 0.66 | 0.78 |
| | | 0.4 | 0.88 | 0.42 | **0.68** | **0.79** |
| Epeka | LatxaLora | - | 0,95 | 0,29 | 0,27 | 0,74 |
| | | 0.3 | 0.9 | 0.38 | 0.57 | 0.78 |
| | | 0.4 | 0.89 | 0.42 | 0.61 | 0.79 |

Taula 9: RoBERTa-euscrawl eta Latxa ereduak diluzio probabilitate ezberdinekin entrenatuta. Emaitzak garapen atalekoak dira.

egin ahala gaindoitu egiten da entrenamenduan eta garapenean okertzen joaten da. Bi datu-multzoetan gaindoitzea gertatzen ari denez domeinuaren eraginaren errua ez dela ondorioztatzen dugu. Entrenamenduak erregularizatzeko diluzio balioarekin jolastu dugu orokortzeko gaitasuna handiagotzeko.

Diluzioak, *drop-out* ingelesez, sailkapen buruko neurona batzuk $p$ probabilitatearekin desaktibatzen ditu, bektoreko dagozkien balioak 0-rekin ordezkatuz. Eragiketa hau 9. eta 10. irudietan $Emb_{CLS}$ eta $Emb_N$ sailkapen buruetan sartu aurretik egiten da. Entrenamenduan zehar bakarrik egiten den eragiketa bat da eta helburua ezaugarri batzuekiko gaindoitzea ekiditea da.

RoBERTa-euscrawl eta Latxa ereduekin egin dira esperimentuak eta 9. taulak erakusten digun bezala, RoBERTa-euscrawl-en kasuan emaitza okerragoak lortu dira, EZ_GAI itzultzeko joera handiagoa baita. Latxak berriz emaitza hobeak lortu ditu, AT orokorra 0.74-tik 0.79-raino igoz. Latxak diluzioarekin duen onura ikusita $p$ balioa 0.4-ra igo dugu hobekuntzarik dagoen antzemateko, baina 0.3 eta 0.4 arteko aldea oso txikia da ia-ia antzeko joera izanik.

Diluzioak datu-multzoetako esperimentuetan duen eragina interesgarria da, oinarri lerroan Epeka entrenatzea egokiagoa dela ikusi dugu, baina diluzioa erabiliz gero, Nahastutako datuekin emaitza orekatuagoak lortzen dira. Alde hau 9. taulako edozein esperimen-

| Banaketa | Eredua | EDA | AT Baxua | AT Duda | AT Altua | AT |
|---|---|---|---|---|---|---|
| Epeka | RoBERTa-euscrawl | ✗ | 0,85 | 0,50 | **0,65** | 0,77 |
|  |  | ✓ | 0,87 | **0,51** | 0,49 | 0,75 |
| Nahastuta | RoBERTa-euscrawl | ✓ | 0.82 | 0.48 | 0.63 | 0.74 |
| Epeka | RoBERTa-euscrawl | SO | 0,84 | 0,45 | 0,57 | 0,75 |
| Nahastuta | RoBERTa-euscrawl | SO | **0,89** | 0,43 | 0,59 | **0,78** |

Taula 10: EDA erabilita egindako esperimentuak. Emaitzak garapen atalekoak dira.

tutan antzeman daiteke.

Erregularizazio atal honen ondorio gisa, diluzioak eragin oso txikia duela RoBERTa ereduetan ikus daiteke, emaitza orekatuak lortzen dira oinarri-lerro gisa erabili dugun ereduarekin. Latxarekin aldiz, emaitza hobeak eta erregularizatuagoak lortzen dira RoBERTa-euscrawlen pareko emaitzak izanik. Latxaren kasuan, diluzioa eta Nahastuta datu-multzoa erabiltzeak emaitza apur bat hobeak lortzen ditu Epeka baino.

### 4.3.2 EDA esperimentuak

3.3.1. atalean azaldu bezala, EDA datu eskasiko ataza eta egoeretan aukera ona izan daiteke sistemen errendimendua hobetzeko. Gure atazan datu ugari ez dagoenez emaitzak hobetzeko teknika egokia izan daiteke.

EDAk 4 heuristiko ezberdin erabiltzen ditu datu gehikuntza egiteko eta 2 hiperparametro ditu. $\alpha$ heuristiko bakoitza aplikatzeko probabilitatea eta $n$ esaldi bakoitzeko sortuko diren esaldi sintetiko berriak. Hiperparametro bilaketa egitea kostu handikoa da EDAren kasuan $n$ handitu ahala entrenamendu instantzia gehiago direlako. Kostuak murriztearren, $\alpha = 0.2$ balioa zehaztu dugu 4 heuristikoentzat eta n berriz 7, horrela instantzia kopurua 8 bider handiagotzen delarik.

Lehen esperimentu honen emaitzak 10. taulako goikaldean ikus ditzakegu eta EDA erabilita edo erabili gabe emaitza antzekoak lortzen dira. Hala ere, orekatuena oinarri-lerro gisa dugun Epeka entrenatutako RoBERTa-euscrawl eredua dugu. Arrazoiren bat topatzearren jarraiko hipotesia osatu dugu: EDAren 4 algoritmoetatik 3k (ATx, AT eta AE) esaldian aldaketa handiak egiten dituzte eta gure atazan GAI izatetik EZ_GAI izateraino alda dezake esaldi bat. Adibidez, hurrengo esaldi hau ongi idatzita dago, hau da, GAI izango litzateke: "Atzo mendira joan nintzen eta bero egiten zuen"3 algoritmo horiek erabilita esaldi hau lor genezake: "Atzo mendira joan nintzen eta bero bide egin egiten zuen". Esaldi sintetiko hau ez dago ongi idatzita, baina jatorrizkoaren etiketa mantentzen denez entrenamendua nahasgarria izan daiteke ereduarentzat.

Azaldutako arazoa ekiditearren, EDAren SO heuristikoa (sinonimoen ordezkapena) bakarrik erabiliko dugu lehengo hiperparametroak mantenduz. Horrela, 10. taulako beheko esperimentuak egin ditugu eta oraingoan ere emaitzak oso antzekoak dira. EDA erabilita ez da onurarik lortzen gure testuinguruan eta kostu konputazionala igotzen denez ez da teknika egokia gure sistema garatzeko.

| Banaketa | Eredua | SCL | AT Baxua | AT Duda | AT Altua | AT |
|---|---|---|---|---|---|---|
| Nahastuta | RoBERTa-euscrawl | ✓ | 0,84 | 0,43 | **0,67** | 0,76 |
| Epeka | RoBERTa-euscrawl | ✓ | 0,85 | 0,46 | 0,63 | 0,77 |
|  |  | ✗ | **0,85** | **0,50** | 0,65 | **0,77** |

Taula 11: SCL galera funtzioa erabilita egindako esperimentuak. Emaitzak garapen atalekoak dira.

Eskuartean dugun ataza hizkuntzaren ezaugarri sakonekin lotuta dago. EDAren algoritmoek esaldien zentzua alda dezakete eta kasu batzuetan esaldia ulertezin bilakatu, honek testuaren zuzentasuna eta koherentzian eragin zuzena izan lezake. Sortutako esaldi berriek jatorrizko esaldiaren klasea mantentzen dutenez, sistemak seinale nahasgarriak jaso eta entrenamendua oztopatu lezake. Aberastasunean ere eragin sakona du SO funtzioak bereziki, sinonimo gehiegi erabiltzen badira, testuaren aberastasuna alda lezake, jatorriz aberastasun gutxiko esaldi bat oso aberats bilakatuz.

### 4.3.3 SCL esperimentuak

SCL galera balioak embedding espazioan etiketa bera duten instantziak gerturatu eta ezberdinak dituztenak urrundu egiten ditu. Galera balio interesagarria da gure atazan, GAI eta EZ_GAI diren instantzien arteko distantzia handitzen bada sistemak errazago sailkatu beharko lituzkeelako dudako kasuak.

Hipotesia egiaztatzeko RoBERTa-euscrawl eredu bat SCL eta CE galera funtzioen batazbestekoa eginda entrenatu dugu, $\lambda$ balioa 0.5 jarrita eta $\tau$ 0.2. Esperimentu honen emaitzak 11. taulan ikus ditzakegu eta orokorrean SCL erabiliz gero ez dira emaitzak hobetzen, oso antzeko portaera erakutsi du ereduak. Dudako kasuak okerrago sailkatzen dituzte eta GAI eta EZ_GAI direnak berriz oso antzeko (puntu bat gorabehera)

Esperimentu hauen emaitza ez desiragarriak ikusita, honakoa ondorioztatu dugu. Gure kasuan klase bakoitzeko testu guztiak ez dute nota berdina, beraz 0 eta 30 duten instantziak oso urrun egon beharko lirateke eta 14 eta 15 direnak berriz oso gertu, baina galera funtzio honek ez du hau kontuan hartzen eta bi egoeretan berdin-berdin jokatzen du. Ondorioz, gure instantziak oker urrunduz emaitzak ez dira hobetzen eta SCL galera funtzio ahaltsuago bat beharko genuke notaren arabera urrunketa hori lehuntzeko.

### 4.3.4 Ereduen aukeraketak

Esperimentu ezberdinetako ereduak garapeneko emaitzen arabera aukeratu ditugu. Aukeraketa horiek egin ondoren, test eta barne-test banaketetan konparatuko ditugu eredurik egokiena zein den zehazteko.

HE aukeraketa eta 3 hobetzeko esperimentu multzoetatik ederurik onenak konparatuko ditugu, honakoak izanik: RoBERTa-euscrawl Epeka eta Nahastuta (4.2), LatxaLora Epeka eta Nahastuta diluzio-p 0.4(4.3.1), RoBERTa-euscrawl Nahastuta EDA-SO (4.3.2) eta RoBERTa-euscrawl SCL Nahastuta eta Epeka.

Sailkatzaile buru bat erabiltzeaz gain, Latxa ereduak adibide-urriko aginduen bidez ebaluatu ditugu. Latxa ereduak instrukzioak jarraitzeko entrenatu gabe daudenez, LM Evaluation Harness liburutegia (Sutawika et al., 2023) erabili dugu ebaluazioak egiteko. Latxaren sarrera token kopuru maximoa eta testuen luzera kontuan hartuta, adibide gisa 3 testu sartzen zaizkio eta ondoren ebaluatu behar duena. C. apendizean ikus daiteke proba hauek egiteko erabili den sarreraren adibidea, Baxua, Dudako eta Altua diren testuek osatzen dute; azpitalde bakoitzeko testu bat aukeratu da. Harness liburutegiarekin 3 eredu tamainak (7B, 13B eta 70B parametroko ereduak) probatu ditugu, konputazio kostua ez baita horren handia.

12. tauletan ereduen test eta barne-test emaitzak ikus ditzakegu. Emaitzak ikusita, AT balio altuena LatxaLora Nahastuta diluzio-p 0.4 ereduak ditu, bai Test baita barne-test banaketetan. Altua testuen kasuan, Testean RoBERTa-euscrawl Nahastuta SCL eta barne-testean RoBERTa-euscrawl Epeka da. Dudakoen kasuan, bi baneketetan RoBERTa-euscrawl Nahastuta SCL. Baxua testuetan berriz, RoBERTa-euscrawl Nahastuta. Latxa ereduek adibide urriko ebaluazioetan lortzen dituzten emaitzak oso oso eskasak dira, gainera 70B-ko ereduak 7 eta 13B-ekoek baina emaitza okerragoak lortzen ditu.

Emaitzak oso antzekoak dira eta ez da erraza eredu, teknika eta datu-multzo egokienak zein diren zehaztea. Dirudienez Test eta barne-test multzoetan Nahastuta entrenamenduak emaitza hobeagoak ematen ditu orokorrean. HEen kasuan, asmatze-tasa orokor altuena Latxa ereduek dute, baina RoBERTak orekatuagoak izan daitezkeela erakutsi dute. EDAk ez duela eredu hobeagorik sortzen ikus dezakegu eta SCLk Nahastuta datu multzoan oso eredu orekatuak sortzen dituela.

Barne testaren asmatze-tasak altuagoak izatea espero zen, baina ikusi da domeinuko eta domeinuz kanpoko ebaluazioek ez dutela eragin handirik. Baliteke gure atazean domeinuaren eragina txikia izatea.

5. atalean zehatzago aztertuko dira eredu hauek, errendimendua sakonago aztertuz.

### 4.4 Datu kopuruaren eragina

Datu sintetikoekin egindako gehikuntzak onurarik ez dakarrela ikusita 4.3.2, datu gehiago erabiltzeak duen onura aztertu nahi izan dugu. Horretarako azpi datu-multzoak osatu ditugu eta esperimentu ezberdinak egin ditugu datu gehiego erabiltzearen eragina aztertzeko.

**Epeka datu-multzoa**   22Api zatiko 4000 testuk osatzen dute, beraz, modu inkrementalean datu gehiago erabilita entrenamenduan duen eragina neurtu nahi izan dugu. Esperimentu hauen emaitzak 13. taulan ikus ditzakegu eta datu gehiago erabilita onura dagoela ikusten da, gehien bat beti gehien agertzen den klasea (EZ_GAI) aurreesateko joetan.

**Nahastuta datu-multzoa**   22Api-ko 4000 testu eta 22Urr-ko 2000 testuk osatzen dute. Datu-multzo honetan proba ezberdinak egin dira 22Api eta 22Urr testu kopuruekin, horrela domeinu berriko testuak erabiltzearen eragina azter dezakegu eta ez bakarrik testu kopuruarekin sistemaren errendimenduaren hobekuntza. 14. taulan agertzen dira esperimentu

| Banaketa | Ereduak | AT Baxua | AT Duda | AT Altua | AT |
|---|---|---|---|---|---|
| Nahastuta | RoBERTa-euscrawl | 0.84 | 0.45 | 0.73 | 0.77 |
| | + EDA-SO | 0.76 | 0.63 | 0.74 | 0.74 |
| | + SCL | 0.71 | 0.66 | 0.87 | 0.75 |
| | LatxaLora + diluzio-p 0.4 | **0.85** | 0.47 | 0.77 | **0.79** |
| Epeka | RoBERTa-euscrawl | 0.75 | 0.59 | 0.81 | 0.75 |
| | + SCL | 0.72 | 0.64 | 0.83 | 0.75 |
| | LatxaLora + diluzio-p 0.4 | 0.80 | 0.50 | 0.75 | 0.76 |
| 3-shot | Latxa-7 | 0.09 | 0.75 | 0.86 | 0.41 |
| | Latxa-13b | 0.29 | 0.59 | 0.69 | 0.44 |
| | Latxa-70b | 0.00 | **0.88** | **1.00** | 0.40 |

(a) Test emaitzak

| Banaketa | Ereduak | AT Baxua | AT Duda | AT Altua | AT |
|---|---|---|---|---|---|
| Nahastuta | RoBERTa-euscrawl | **0.90** | 0.39 | 0.55 | 0.75 |
| | + EDA-SO | 0.86 | 0.40 | 0.51 | 0.71 |
| | + SCL | 0.75 | 0.59 | 0.72 | 0.72 |
| | LatxaLora + diluzio-p 0.4 | 0.83 | 0.49 | 0.73 | **0.76** |
| Epeka | RoBERTa-euscrawl | 0.78 | 0.54 | 0.78 | 0.75 |
| | + SCL | 0.78 | 0.52 | 0.74 | 0.74 |
| | LatxaLora + diluzio-p 0.4 | 0.84 | 0.47 | 0.68 | 0.75 |
| 3-shot | Latxa-7 | 0.15 | 0.76 | 0.75 | 0.51 |
| | Latxa-13b | 0.32 | 0.65 | 0.82 | 0.45 |
| | Latxa-70b | 0.00 | **0.81** | **0.98** | 0.38 |

(b) Barne test emaitzak

Taula 12: Esperimentu multzo bakoitzeko eredu onenen test emaitzak hauen emaitzak. Esperimentu hauen arabera, 22Urr 1000 bakarrik erabilita entrenatzea da orekatuena, Dudazko eta Altua diren testuak egokien sailkatzen dituelako. Hala ere, esperimentu ezberdinen artean dagoen aldea ez da oso handia eta zenbaki absolutuetan egokiagoa da 22Api 4000 + 22Urr 1000 erabiltzea.

Esperimentu hauetatik ezin da ondorio garbirik atera, badirudi Epeka zatian hobe dela testu kopuru ia guztia erabiltzea. Nahastutako datu-multzoan berriz ikusi da 22Api eta 22Urr testuen nahasketak sistema ez-orekatua sortzen duela.

| Data Kop. | AT Baxua | AT Duda | AT Altua | AT |
|---|---|---|---|---|
| 22Api 1000 | **0.95** | 0.31 | 0.34 | 0.75 |
| 22Api 2000 | 0.88 | 0.46 | 0.49 | 0.76 |
| 22Api 3000 | 0.82 | **0.46** | **0.68** | 0.75 |
| 22Api 4000 | 0.84 | 0.42 | 0.65 | **0.76** |

Taula 13: Epeka datu-multzoaren probak datu kopuru ezberdinekin. Emaitzak garapen atalekoak dira.

| Data Kop. | AT Baxua | AT Duda | AT Altua | AT |
|---|---|---|---|---|
| 22Urr 1000 | 0.8 | **0.46** | **0.67** | 0.74 |
| 22Urr 2000 | 0.86 | 0.40 | 0.63 | 0.77 |
| 22Api 1000 + 22Urr 2000 | 0.87 | 0.43 | 0.64 | 0.78 |
| 22Api 2000 + 22Urr 2000 | 0.87 | 0.40 | 0.65 | 0.78 |
| 22Api 3000 + 22Urr 2000 | 0.88 | 0.39 | 0.56 | 0.76 |
| 22Api 4000 + 22Urr 2000 | **0,95** | 0,33 | 0,47 | 0,78 |
| 22Api 4000 + 22Urr 1000 | **0,95** | 0,34 | 0,49 | **0,79** |

Taula 14: Nahastuta datu-multzoaren probak datu kopuru ezberdinekin. Emaitzak garapen atalekoak dira.

# 5 Analisia

4. atalean hainbat esperimentu eta haien emaitzak azaldu ditugu. Orokorrean badirudi eredu ia guztien portaera oso antzekoa dela, asmatze-tasan gorabehera gutxi egonik. Eredu batzuk joera handia dute beti EZ_GAI esateko, beste batzuk orekatuagoak dira. Domeinuaren eragina ere oso txikia dela ikusi dugu 12a. atalean.

Ereduen portaera eta egiten diren akatsen arrazoia argitzeko, 2 azterketa mota diseinatu eta burutu ditugu.

## 5.1 Kalibrazio azterketa

Gure hurbilpena gehien bat sailkapen bitarrean oinarritu da, hau da, testuak EZ_GAI edo GAI diren aurresatean. Hala ere, testu bakoitzak 0-30 tartean lortutako nota eskuragarri dugunez, azterketa sakonago bat egin dezakegu.

Intuizioa jarraituz testuak zenbat eta 15 (gainditzeko muga) puntutik gertu egon are eta zailagoak izan behar dira sailkatzeko, hau da, ereduek etiketa bakoitzari ezartzen dion softmax (sasi-probabilitatea) balioa txikiagoa izan beharko da. Honi ereduen kalibrazioa deitzen zaio eta sailkapen ereduen gaitasunak aztertzeko ohikoa da.

Gure ereduen kalibrazioak neurtzeko entrenatu direnetatik errendimendu onena izan dutenak aukeratu dira, 4.3.4. atalean aztertutakoak alegia.

20. eta 21. irudietan hainbat kalibrazio grafika ikus ditzakegu. Lehenik eta behin ongi ulertu behar dugu grafikak erakusten diguna. X ardatzak azterketa bakoitzaren eskuzko nota adierazten digu, Y ardatzak azterketa bakoitzari sailkatzaileak esleitzen dion konfidantza balioa (1 eta -1 tartera mapatuta) adierazten du eta azkenik koloreak testuaren jatorrizko etiketa adierazten du (laranja GAI eta urdina EZ_GAI direnentzat). Perfektuki kalibratutako eredu batek diagonal bat marraztu beharko luke, hau da, notak gora egin ahala azterketari esleitutako probabilitatea altuagoa izan beharko luke.

Kalibrazio grafikak aztertuta hainbat ondorio atera ditzakegu eredu hauen portaeretatik:

- RoBERTa-euscrawl Epeka eta Nahastuta alderatuta, lehenak joera oso handia du etiketa bakoitza probabilitate masa oso handi batekin esleitzeko. Horregatik puntu gehienak goikaldean edo behekaldean daude eta oso gutxi dira erdiko zonaldean daudenak. Nahastutako entrenamenduaren ondorioz, probabilitate masa horiek muturreko balioetatik urruntzen dira, erdiko zonaldean puntu gehiago egonik eta eredu kalibratuago bat izanik.

- EDArekin entrenatuta RoBERTa-euscrawl ereduak probabilitate masa oso txikiak esleitzen ditu, horregatik testu guztiak y=0-ren inguruan kokatzen dira. EDA teknika egokia ez dela ondorioztatu dugu 4.3.2. atalean eta kalibrazio grafika honek ideia bera berresten du.

- SCL bidezko entrenamenduek grafika oso antzekoak lortzen dituzte. RoBERTa-euscrawl Epeka ereduaren oso antzeko kalibrazioa dutela ikusten da eta probabilitate oso altua esleitzen dietela testuei ikus dezakegu.

- Latxa ereduen kasuan kalibrazio oso txarra ez dutela ikus daiteke. RoBERTa-euscrawl Nahastuta ereduaren antzeko portaera dutela erakutsi dute. Badirudi eredu hauek zailtasunak dituztela 15 inguruko testuak sailkatzean, hau esperotakoa da horiek baitira dudazko kasuak.

Kalibrazio grafikak modu oso bisual batean ereduaren portaera aztertzen laguntzen dute, baina, zenbakizko metrikarik gabe oso zaila da eredu kalibratuena zein den ondorioztatzea. Adibidez, Latxaren kasua ezin daiteke begi bistaz Epeka edo Nahastuta entrenatzea egokiagoa den argitu. Metrika objektiborik gabe grafikekin konparaketak egitea subjektiboa da.

Subjektibitatearen arazoa konpontzearren aurresandako etiketak eta egiazko etiketen arteko pearson korrelazioa kalkulatu dugu. 15. taulak eredu bakoitzaren korrelazio balioa erakusten digu. Grafiketan ikusten zena zenbakizko metrika honek ere erakusten digu, EDAk entrenamenduko kalibrazioa okertzen du, SCL ereduek oso antzeko joera dute eta RoBERTa-euscrawl ereduen kasuan Nahastuta datu-multzoan entrenatutakoak kalibrazio hobea du. Azken hau ere Latxaren kasuan ikus dezakegu, korrelazio hobea du Nahastuta entrenatu den ereduak. Korrelazioaren arabera eredu kalibratuena Latxa eredua da Nahastuta entrenamenduan eta diluzio-p 0.4 duena.

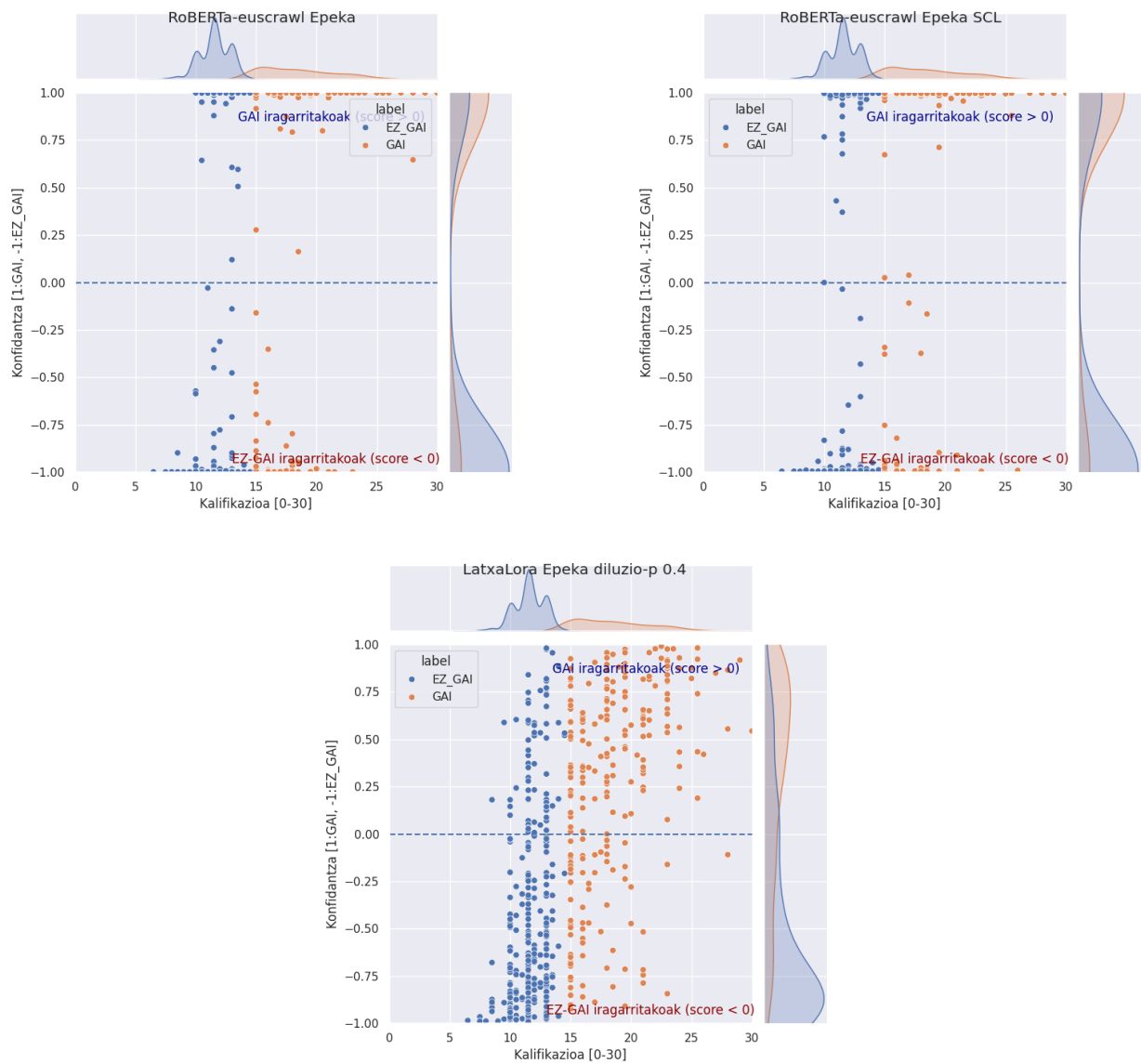

Irudia 20: Epeka entrenatutako eredu ezberdinen test ataleko kalibrazioa.

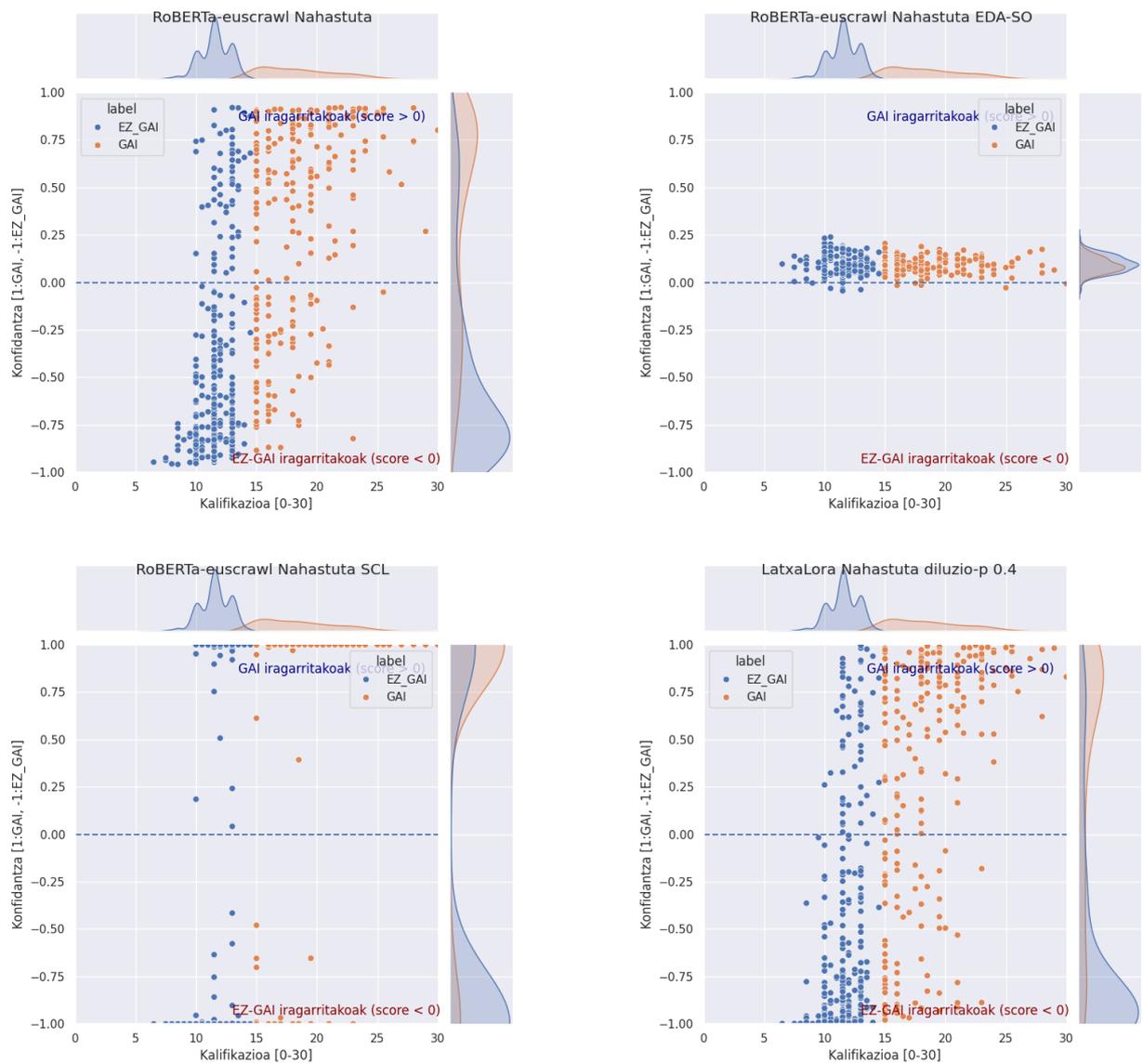

Irudia 21: Nahastuta entrenatutako eredu ezberdinen test ataleko kalibrazioa.

| Banaketa | Eredua | Pearson korrelazioa |
|---|---|---|
| Epeka | RoBERTa-euscrawl | 0.5502 |
| | + SCL | 0.5394 |
| | LatxaLora + diluzio-p 0.4 | 0.6385 |
| Nahastuta | RoBERTa-euscrawl | 0.6709 |
| | + EDA-SO | -0.0695 |
| | + SCL | 0.5690 |
| | LatxaLora + diluzio-p 0.4 | **0.6713** |

Taula 15: Ereduen pearson korrelazioa

## 5.2 Artefaktuen azterketa

Ereduen entrenamenduen emaitzetan, entrenamenduko testuak ia perfektuki sailkatzeko gai direla ikus daiteke 18. taulan. Hala ere, garapen edota test banaketetan emaitzak okerragoak dira eta honek entrenamenduan gaindoitze bat dagoela adierazten digu.

Gaindoitzearen ondorioetako bat ereduak artefaktuak edota nahi ez diren ezaugarriak ikasten dituztela izan daiteke. Gerta liteke entrenamenduan zehar ereduak orokortzeko behar duena ez ikasi eta beste ezaugarri batzutan zentratu izana.

Arazo hau identifikatzea ez da tribiala eta oso zaila izan daiteke. HE neuronalak kutxa beltzak direnez, oso zaila da atentzioa zein ezaugarritan zentratzen duten jakitea. Hainbat saiakera egin dira trasnformers arkitekturako atentzio balioak iterpretatzen ereduen arrazoiketa azaleratzeko (Hao et al., 2021) eta orokortzeko gaitasuna areagotzeko (Moon et al., 2021). Teknika hauek interesgarriak izanda ere, bere mugak dituzte eta ez dute arrazoiketa sakonik azaleratzen.

Gure kasuan badakigu ereduak ikasi beharreko ezaugarri linguistikoak zein diren: Zuzentasuna, Aberastasuna, Koherentzia, Kohesioa eta Egokitasuna. Hau kontutan hartuta analisi sinple bat diseinatu dugu, artefaktuak ikasten dituen edo ez zehazteko. Jarraiko pausuak jarraitu dira analisian:

1. Erabiliko diren testu multzotik GAI direnak bakarrik hartu.

2. Ebaluatu nahi den sailkatzaileari etiketa aurresateko eskatu.

3. Asmatu dituen testuak hartu (GAI direla aurresan dituenak bakarrik) eta testuen esaldiak ausaz berrordenatu. Teorian Kohesio eta Koherentzia ezagatik EZ_GAI bihurtu beharko lirateke.

4. Ereduari berrordenatutako testuak eman eta EZ_GAI bezala aurresan dituenen protzentaia kalkulatu.

Algoritmoaren ideaia oso sinplea da, baina esaldiak berrodenatuz koherentzia eta kohesio balioak oso txarrak izan beharko lirateke. Gainera, Egokitasuna eta Zuzentasunak ere eragin oso negatiboa izan beharko lukete, lehenengoaren kasuan testuaren forma guztiz desitxuratzen baita eta bigarrengoaren kasuan erakusle, pertsonai eta abar guztiz aldatzen direlako. Ereduak berrordenatutako testuak GAI bezala sailkatzen jarraitzen badu, artefaktuak edo nahi ez diren ezaugarriak ikasten dituelako izango da, zenbat eta EZ_GAI gehiago aurresan, ereduaren orokortzeko gaitasunak hobea izan beharko luke.

16. taulak eredu bakoitzak berritsuratutako testuak EZ_GAI gisa aurresateko duen joera erakusten digu, zenbat eta balio altuagoa hobe. Datu-multzoetako banaketa guztiak erabili dira testu ezberdinak egokiago konparatzeko. Emaitzak erakusten digutenaren arabera, LatxaLora diluzio-p 0.4 Epeka eta Nahastuta ereduek artefaktu edo nahi ez diren ezaugarriei jaramon gutxiago egiten diete.

16. taulak entrenamendu metodo bakoitzaren analisi sakonago bat egitea ahalbideratzen digu. RoBERTa-euscrawl oinarrizko ereduetan Nahastuta entrenamendua eginda entrenamenduko joera asko hobetzen da, honek Epeka entrenamenduan memorizazio bat egon dela adieraz dezake, hau da, ereduak testuak buruz ikasteko joera duela. EDA-SO entrenamenduan artefaktu gehiegi ikasten direla ikus dezakegu, ia banaketa guztietan balio oso baxuak lortu dituelako. SCL eta Epeka entrenatzeak onura dakarrela ikus dezakegu oinarrizko RoBERTa ereduekin alderatuta. Epeka entrenamenduak emaitza hobeak eman ditu orokorrean, hau gertaera bitxia da, Nahastuta entrenamenduan testu gehiago eta gai ezberdinetakoak daudenez orokortzeko gaitasun hobea espero baita.

Azkenik, test zatian eredu guztiek emaitza baxuagoak lortu dituzte barne-test edota garapen zatiarekin konparatuz. Honek test zatiko testuak gainontzekoak baino apur bat ezberdinak direla erakusten digu, azken finean ez baitira automatikoki transkribatuak izan.

| Banaketa | Eredua | Train | Garapen | Test | Barne-test |
|---|---|---|---|---|---|
| Epeka | RoBERTa-euscrawl | 0.0492 | 0.2717 | 0.1176 | 0.2939 |
| | + SCL | 0.0935 | 0.3954 | 0.1856 | 0.3333 |
| | LatxaLora + diluzio-p 0.4 | **0.4684** | **0.5364** | **0.2791** | **0.4259** |
| Nahastuta | RoBERTa-euscrawl | 0.2628 | 0.2708 | 0.0854 | 0.2614 |
| | + EDA-SO | 0.0 | 0.0465 | 0.0072 | 0.1384 |
| | + SCL | 0.0425 | 0.2627 | 0.1610 | 0.2303 |
| | LatxaLora + diluzio-p 0.4 | 0.4574 | 0.4389 | 0.2356 | 0.3772 |

Taula 16: Eredu bakoitzak berrordenatutako esaldiak EZ_GAI gisa aurresateko joera.

# 6  Ondorioak eta etorkizuneko lanak

Proiektuan zehar ataza konplexu hau ebazten saiatu gara eta lehen urbilpen batzuk probatu ditugu sailkatzaile on bat diseinatu nahian. Orokorrean oinarri-lerroa ez da hobetu, honek atazaren konplexutasuna adierazten digu eta entrenamendurako testu urritasuna. Jarraian atalka ateratako ondorioak azalduko ditugu.

## 6.1 Datu-multzoen eragina

Esperimentu guztiak bi datu multzo erabilita egin dira, Epeka eta Nahastuta, azken honek bi azterketa epealdi ezberdinetako testuak nahasten dituelarik. RoBERTa oinarrizko ereduetan, Nahastuta entrenamenduak beti EZ_GAI itzultzeko joera handitzen duela dirudi, Epeka entrenatutakoak asmatze-tasa orekatuagoa duela erakutsi baitu.

SCL eta EDA esperimentuetan emaitza oso antzekoak lortzen dituzte bi multzoek. Ondorioz, ezin da ondorio garbirik atera bi esperimentu hauetan datu-multzoak duen eraginean. Latxaren kasuan Nahastuta datu-multzoak emaitzak hobetzen ditu, bai asmatze-tasa, baita 5. ataletako azterketetan.

Oso zaila da entrenamenduan epealdi ezberdinetako testuak edukitzeak dakarren onura aztertzea. Ondorio garbirik ezin dugu atera eta etorkizuneko lanetarako geratuko da zalantza hau argitzea. Epealdi ezberdinetako transkripzio gehiago izanda datu-multzoen konparaketa egokiagoa egin genezake eta ondorio sakonagoak atera.

## 6.2 Esperimentuen ondorioak

Oinarrizko HEen aukeraketa egitean, 4.2. atalean, RoBERTa-euscrawl kodetzailea hartu dugu oinarri-lerro gisa.

Latxaren erregularizazio eza kontuan hartuta, entrenamenduak erregularizatzeko diluzio-p balioarekin jolastu dugu. RoBERTa-euscrawl-en kasuan diluzio probabilitateak igotzeak ez du inongo hobekuntzarik suposatu, baina Latxa ereduetan berriz, emaitzak asko hobetu dira. 5. ataleko azterketetan erregularizatutako Latxak gaitasun hobeak erakutsi ditu. Hala ere, asmatze-tasa orekatuagoa mantendu da RoBERTa-euscrawl Epeka entrenamenduan.

Erregularizazioaz gain, EDA eta SCL teknikak erabilta probak egin dira. EDAren kasuan asmatze-tasak ez dira hobetu eta analisiak eginda errendimendu eskaxa eta desira ez dena erakutsi du. SCLren kasuan, asmatze-tasak ez dira hobetu, kalibrazio aldetik ere emaitzak ez dira oso onak 5.1, baina 5.2. ataleko artefaktu azterketan oinarrizko RoBERTa ereduak baino emaitza egokiagoak lortzen ditu.

Orokorrean oinarri-lerroa hobetzeko egin diren esperimentuek ez dute onura gehiegi suposatu, EDAren kasuan adibidez, kostu konputazionala ere handitzen da. Asmatze-tasaren arabera oinarri-lerroa da eredu egokiena, baina, 5. ataleko azterketek Latxa ereduak egokiagoak direla ondorioztatzen dute.

Txostenan agertu ez arren, erregresio bitartezko hurbilpenak ere landu dira, emaitza oso antzekoak lortzen dira sailkapen edo erregresio bezala ebatzita, hobekuntza arbuiagarria da eta aurresandako noten eta egiazko noten arteko korrelazioa eskasa da. Onurarik ez dagoela ikusita sailkapen bidearekin jarraitzea erabaki da.

## 6.3 Datu kopuruen ondorioak

4.4. atalean datu kopuru ezberdinak erabilita ereduengan dagoen eragina aztertu da. EDA esperimentuetan emaitzak hobetzen ez zirela ikusita azterketa hau garrantzitsua zen, agian testu gehiegi erabilita emaitzak okertzen zirelako.

Hala ere, esperimentuak eginda, 22Api testuetan guztiak erabiltzea hobea dela ikusi da. Baina, 22Urr testuekin gauza oso bitxi bat ikusi dugu, zenbat eta testu gehiago erabilita eta 22Api-rekin nahastuz emaitza orekatu batetik beti EZ_GAI itzultzeko joera batera igaro da. Hau oso bitxia da eta datu-multzo ezberdinek eragiten duten joerak sakonki aztertu beharko dira transkribatutako testu gehiago ditugunean.

## 6.4 Analisien ondorioak

5. ataleko bi azterketen ondorioz, honako ondorioak atera ditugu:

1. Eredu gehienak sailkatzerako orduan etiketa bakoitzari probabilitate handiena esleitzen dio. Latxa eta RoBERTa-euscrawl Nahastuta ereduak izan ezik, baina, azken hauek ere zalantza handiak izaten dituzte testuekin eta ez daude guztiz ongi kalibratuta.

2. Sailkatzerako orduan ereduek ez dituzte nahi ditugun ezaugarriak aztertzen, 5.2. atalean ikus daitekeen bezala. Joera hau nabariagoa da RoBERTa-euscrawl ereduetan. Latxa ereduak egokiagoak direla ikusi da azterketa horretan, baina, oraindik urrun daude emaitza onetatik.

Kalibrazio azterketetan eta artefaktu azterketetan lortutako emaitzak korrelazioren bat dutela dirudi, bietan Latxa izan da eredurik egokiena, EDAk emaitza txarrak eman ditu eta oinarrizko RoBERTa-euscrawl ereduetan Nahastuta entrenamenduak portaera hobetzen dio ereduari.

## 6.5 Etorkizunerako lanak

Proiektuan zehar ez dugu nahi adina esperimentu eta azterketa egiteko denbora izan. Gainera, testuen transkripzioa martxan ari dela ez dugu aukerarik izan corpus guztiarekin lan egiteko, beraz, testu gehiago izanik emaitzak eta ondorioak alda litezke.

HEei buruz, euskararako dagoen eredu kodetzaile berri eta onena zaharkituta geratu da, ingeleserako arkitektura berriak eta hobekuntzak aurkeztu dira, deBERTa (He et al., 2021) ereduak adibidez eta interesgarria izango litzateke gaitasun handiagoko HE berri bat edukitzea.

Probatzeko geratu dira gaindoitzeak ekiditeko beste hainbat teknika. Adibidez, izen bereziak eta izen ondoak maskaratzeak ereduak orokortzeko gaitasuna izateko aukera zabal dezake. Baita ere, Moon et al. (2021)-ek aurkeztutako maskaratze estrategia interesgarria izan daiteke, orokortzeko eta artefaktuak ekiditeko.

Bestalde, ataza sailkapen bitarra beharrean beste modu batera ebatzi liteke. Azterketa bakoitzeko 5 irizpideen notak eskura ditugu, beraz, 5 eredu independiente entrena genitzake bakoitza irizpide bakoitza ikasirik eta ondoren azken nota kalkulatu. Horrela, agian ereduek errazago ikasiko lukete behar dituzten ezaugarriak ebaluatzen.

Azkenik, Latxa eredu sortzailearekin proba gehiago egitea falta izan zaigu. Proiektuan zehar 7b modeloarekin egin ditugu entrenamenduak eta azterketak, baina transkribatutako testu gehiago izanda eredurik ahaltsuenarekin probak eginda artefaktuen azterketetan ikusitako orokortzeko ezaugarriak areagotuko direla espero dugu, eredu ahaltsuago bat sortzeko aukera izanik. Bestalde, proiektu honetan eredu hau kodetzaile moduan erabili dugu eta ez ditugu bere sortzaile gaitasunak ikertu, adibide edota ebaluatzaileek erabiltzen dituzten errubrikak erabil genitzake ereduari argibideak emateko.

# Erreferentziak

# A  Esperimentuetako hiperparametroak

Jarraian esperimentu ezberdinetan erabilitako hiperparametroen taulak. Entrenamenduko datuen banaketek ez dute hiperparametroetan eraginik, hau da, Epeka entrenatutako RoBERTa-euscrawl ereduak eta Nahastuta entrenatutakoak hiperparametro berdinak erabiltzen dituzte.

HAP masterra

| Hiperparametroak | Kodetzaileak* | LatxaLora |
|---|---|---|
| Epoch | 10 | 10 |
| Learning Rate | 5e-5 | 5e-5 |
| Weight Decay | 1e-4 | 1e-4 |
| Warmup | 20 | 20 |
| Scheduler | Cosine | Cosine |
| Batch size | 32 | 4 |
| Gradient Accumulation | 1 | 2 |
| Optimizatzailea | AdamW | AdamW |
| Max length | 512 | 1024 |
| Doitasuna | FP16 | BF16 |
| Atentzioa | Arrunta | Flash attention 2 |
| Lora r/$\alpha$/drop/bias | - | 8/8/0.1/none |
| Diluzio-p† | 0.1 | 0.0 |
| SCL $\lambda/\tau$ † | 0.3/0.1 | - |
| EDA n / $\alpha$ † | 8 / 0.2 | - |

Taula 17: Entrenamendu ezberdinetan erabilitako hiperparametroak. * Kodetzaile guztiak, BERTeus, XLM-RoBERTa eta RoBERTa-euscrawl hiperparametro berdinekin entrenatu ditugu. †hiperparametro hauek dagozkien esperimentuetan bakarrik erabili dira.

# B  Esperimentu guztien emaitza taulak

Proiektuan zehar egindako esperimentu guztien emaitzak 4 tauletan banatu dira. Entrenamenduko emaitzak 18, garapeneko emaitzak 19, testeko emaitzak 20 eta barne-test emaitza guztiak 21

# C  Testuen adibideak

## C.1  Baxua testu baten adibidea

Telemedikuntza abian jarri zen salbuespen egoerari erantzuteko, baina Eusko Jaurlaritzak esan zuen telefonoz eta bideoz osasun arreta ematen jarraituko duela. Izan ere, metodo honek onura asko ekarri dizkigu gure bizimoduaren arabera, baina erabilera metodoa egokia ez bada, alderdi negatibo asko aurki ditzakegu. Lehenik eta behin, gaur egungo bizimoduak ez du zerikusirik duela 30 urte bizi ginenarekin. Gaur egun, gaueko orduak eta teknologia berriek 24 orduz lan egiteko aukera ematen digute, eta, beraz, askotan egunean bizpahiru minutu libre izaten ditugu eta medikuarekin hitz egiteko erabiltzen ditugu, baina ezinezkoa da denbora horretan egitea. medikuarengana joan eta telemedikuntzak nahi duguna egiten laguntzen digu lana utzi gabe. Horrek arriskuan jar dezake, gogoratu medikuek ez zaituztela ikusten eta ikusten badituzu, kalitate eskasa izango da eta ezingo

| Banaketa | Eredua | AT Baxua | AT Duda | AT Altua | AT |
|---|---|---|---|---|---|
| Epeka | Majority Class | 1.00 | 0.16 | 0.00 | 0.36 |
| | BERTeus | 0.98 | 0.86 | 0.92 | 0.95 |
| | XLM-RoBERTa | 1.00 | 0.99 | 1.00 | 1.00 |
| | RoBERTa-euscrawl | 1.00 | 1.00 | 1.00 | 1.00 |
| | + EDA | 1.00 | 1.00 | 1.00 | 1.00 |
| | + EDA-SO | 0.99 | 0.93 | 0.96 | 0.98 |
| | + SCL | 1.00 | 1.00 | 1.00 | 1.00 |
| | + diluzio-p 0.3 | 1.00 | 0.99 | 1.00 | 1.00 |
| | LatxaLora | 1,00 | 0,50 | 0,40 | 0,78 |
| | + diluzio-p 0.3 | 0.94 | 0.80 | 0.87 | 0.90 |
| | + diluzio-p 0.4 | 0.87 | 0.61 | 0.74 | 0.81 |
| Nahastuta | Majority Class | 1.00 | 0.16 | 0.00 | 0.66 |
| | BERTeus | 0.96 | 0.94 | 0.97 | 0.96 |
| | XLM-RoBERTa | 1.00 | 0.22 | 0.00 | 0.66 |
| | RoBERTa-euscrawl | 0.99 | 0.78 | 0.86 | 0.93 |
| | + EDA | 0.99 | 0.98 | 0.99 | 0.99 |
| | + EDA-SO | 1.00 | 0.84 | 0.78 | 0.92 |
| | + SCL | 1.00 | 1.00 | 1.00 | 1.00 |
| | + diluzio-p 0.3 | 0.96 | 0.80 | 0.87 | 0.93 |
| | LatxaLora | 0,99 | 0,43 | 0,37 | 0,78 |
| | + diluzio-p 0.3 | 0,92 | 0,75 | 0,84 | 0,88 |
| | + diluzio-p 0.4 | 0,91 | 0,70 | 0,82 | 0,86 |

Taula 18: Entrenamendu ataleko emaitzak

| Banaketa | Eredua | AT Baxua | AT Duda | AT Altua | AT |
|---|---|---|---|---|---|
| Epeka | Majority Class | 1.0 | 0.15 | 0.0 | 0.70 |
| | BERTeus | 0,899 | 0,448 | 0,515 | 0,772 |
| | XLM-RoBERTa | 1,00 | 0,197 | 0,000 | 0,703 |
| | RoBERTa-euscrawl | 0,845 | 0,503 | 0,649 | 0,770 |
| | + EDA | 0,870 | 0,508 | 0,494 | 0,753 |
| | + EDA-SO | 0,841 | 0,448 | 0,575 | 0,746 |
| | + SCL | 0,850 | 0,464 | 0,633 | 0,766 |
| | + diluzio-p 0.3 | 0.909 | 0.365 | 0.582 | 0.787 |
| | LatxaLora | 0,954 | 0,290 | 0,266 | 0,740 |
| | + diluzio-p 0.3 | 0.904 | 0.377 | 0.566 | 0.780 |
| | + diluzio-p 0.4 | 0.895 | 0.421 | 0.606 | 0.787 |
| Nahastuta | Majority Class | 1.0 | 0.15 | 0.0 | 0.70 |
| | BERTeus | 0,82 | 0,51 | 0,70 | 0,76 |
| | XLM-RoBERTa | 1,00 | 0,22 | 0,00 | 0,70 |
| | RoBERTa-euscrawl | 0,95 | 0,33 | 0,47 | 0,78 |
| | + EDA | 0.82 | 0.48 | 0.63 | 0.74 |
| | + EDA-SO | 0,89 | 0,43 | 0,59 | 0,78 |
| | + SCL | 0,84 | 0,43 | 0,67 | 0,76 |
| | + diluzio-p 0.3 | 0.90 | 0.47 | 0.57 | 0.79 |
| | LatxaLora | 0,97 | 0,24 | 0,25 | 0,74 |
| | + diluzio-p 0.3 | 0,87 | 0,43 | 0,66 | 0,78 |
| | + diluzio-p 0.4 | 0,88 | 0,42 | 0,68 | 0,79 |

Taula 19: Garapen ataleko emaitzak

| Banaketa | Eredua | AT Baxua | AT Duda | AT Altua | AT |
|---|---|---|---|---|---|
| Epeka | Majority Class | 1,00 | 0,09 | 0,00 | 0,60 |
| | BERTeus | 0,80 | 0,41 | 0,69 | 0,73 |
| | XLM-RoBERTa | 1,00 | 0,13 | 0,00 | 0,60 |
| | RoBERTa-euscrawl | 0,75 | 0,59 | 0,81 | 0,75 |
| | + EDA | 0,77 | 0,45 | 0,67 | 0,70 |
| | + EDA-SO | 0,74 | 0,61 | 0,77 | 0,73 |
| | + SCL | 0,72 | 0,64 | 0,83 | 0,75 |
| | + diluzio-p 0.3 | 0,78 | 0,50 | 0,80 | 0,75 |
| | LatxaLora | 0,92 | 0,20 | 0,36 | 0,67 |
| | + diluzio-p 0.3 | 0,83 | 0,44 | 0,72 | 0,75 |
| | + diluzio-p 0.4 | 0,80 | 0,50 | 0,75 | 0,76 |
| Nahastuta | Majority Class | 1,00 | 0,09 | 0,00 | 0,60 |
| | BERTeus | 0,63 | 0,75 | 0,86 | 0,72 |
| | XLM-RoBERTa | 1,00 | 0,13 | 0,00 | 0,60 |
| | RoBERTa-euscrawl | 0,84 | 0,45 | 0,73 | 0,77 |
| | + EDA | 0.73 | 0.45 | 0.69 | 0.69 |
| | + EDA-SO | 0,76 | 0,63 | 0,74 | 0,74 |
| | + SCL | 0,71 | 0,66 | 0,87 | 0,75 |
| | + diluzio-p 0.3 | 0.78 | 0.63 | 0.81 | 0.77 |
| | LatxaLora | 0,95 | 0,23 | 0,36 | 0,69 |
| | + diluzio-p 0.3 | 0,86 | 0,45 | 0,72 | 0,77 |
| | + diluzio-p 0.4 | 0,85 | 0,47 | 0,77 | 0,79 |

Taula 20: Test ataleko emaitzak

| Banaketa | Eredua | AT Baxua | AT Duda | AT Altua | AT |
|---|---|---|---|---|---|
| Epeka | Majority Class | 1,00 | 0,13 | 0,00 | 0,66 |
| | BERTeus | 0,86 | 0,37 | 0,63 | 0,74 |
| | XLM-RoBERTa | 1,00 | 0,22 | 0,00 | 0,64 |
| | RoBERTa-euscrawl | 0,78 | 0,55 | 0,78 | 0,75 |
| | + EDA | 0,81 | 0,46 | 0,60 | 0,71 |
| | + EDA-SO | 0,80 | 0,44 | 0,66 | 0,72 |
| | + SCL | 0,78 | 0,52 | 0,74 | 0,74 |
| | + diluzio-p 0.3 | 0,82 | 0,41 | 0,69 | 0,74 |
| | LatxaLora | 0,95 | 0,28 | 0,29 | 0,69 |
| | + diluzio-p 0.3 | 0,83 | 0,48 | 0,64 | 0,74 |
| | + diluzio-p 0.4 | 0,84 | 0,47 | 0,69 | 0,75 |
| Nahastuta | Majority Class | 1,00 | 0,13 | 0,00 | 0,66 |
| | BERTeus | 0,74 | 0,52 | 0,78 | 0,73 |
| | XLM-RoBERTa | 1,00 | 0,22 | 0,00 | 0,64 |
| | RoBERTa-euscrawl | 0,90 | 0,39 | 0,55 | 0,75 |
| | + EDA | 0.79 | 0.47 | 0.62 | 0.71 |
| | + EDA-SO | 0,86 | 0,40 | 0,51 | 0,71 |
| | + SCL | 0,75 | 0,59 | 0,72 | 0,73 |
| | + diluzio-p 0.3 | 0.84 | 0.45 | 0.69 | 0.75 |
| | LatxaLora | 0,953 | 0,25 | 0,27 | 0,69 |
| | + diluzio-p 0.3 | 0,82 | 0,49 | 0,70 | 0,75 |
| | + diluzio-p 0.4 | 0,83 | 0,49 | 0,73 | 0,76 |

Taula 21: Barne-test ataleko emaitzak

duzula diagnostikatu ondo joanez gero. Eta epaiketa hauek ematen dizkigutenean, batzuetan aurkitzen dugu gure buruaz hornitzeko zerbait, eta bizi garen bitartean iraintzen gaituzte hura ezagutzeko. Laburbilduz, telemedikuntza betidanik erabili izan da larrialdiei erantzuteko eta erabili beharko genuke. Ez dugu merkatua argitzeko astirik izan, baina erabiltzea ez da diagnostikorako aukerarik onena. Izan ere, beti aukerarik ez badago, hobe da hura deitzea zalantzan egotea edo arima bilatzea baino. Uste dut ulertu behar dugula gure bizimodua pixkanaka aldatzen ari dela, baina gauza garrantzitsu hauek egin ahal izateko, fax bidez egin behar dugu.

## C.2 Dudako testu baten adibidea

Teknologia modernoak aurrerapen handiak ekarri ditu medikuntzan. Orain dela urte batzuk, 2020an, Covid-19aren pandemiak telemedikuntza areagotzea ekarri zuen, hau da, mediku bati telefonoz galdetzea. Baina telemedikuntza onuragarria edo kaltegarria izan daiteke gizartearentzat? Etorkizuneko sendagaia izan daiteke?

Aurrerapen teknologiko honek hainbat abantaila izan ditzake. Batek ez du etxetik joan behar. Gainera, telemedikuntza edozein unetan erabil daiteke, beraz, inork ez du denborarik galtzen medikuari kontsultatzeko. Eta ez da bakarrik Covid-19 bezalako egoeretan, baita noizbehinkako kasuetan ere, oso baliagarria dela; adibidez, gripe ahula duzunean.

Bestalde, badira alde txarrak ere. Gaur egun, biztanleriaren zati handi bat gero eta gehiago hasten da telemedikuntza erabiltzen, eta egoera horrek jarraitzen badu, etorkizunean, kontsulta gehiago egingo dira telefonoz mediku bati egindako kontsulta fisikoak baino. Agian ez dirudi, baina kontsulta horiek ondorioak izan ditzakete gizartean eta biztanlerian. Jendea ez zen ospitalera joango, ezta medikuarengana ere, eta erizain eta medikuek ez zuten lanik izango. Beraz, horrela balitz, osasun-langileek ez lukete hainbeste diru irabaziko.

Bestalde, telefonoz diagnostikatzea ez da hain erraza eta ideala. Espezialista batek aurrez aurre egon behar du pazientearekin, behar duten arreta emateko. Horrela, diagnostikoa egin ahal izango da eta tratamendua telefonoz eman ahal izango da, gripe arina izateko. Baina ez beste gaixotasun larriagoetarako. Sintomak badituzu, hurrengoan medikuarengana joan beharko zenuke.

Beste abantaila esanguratsu bat adinekoen arreta da. Populazio honek bisitari askok baino arreta mediko gehiago behar du, beraz, metodo hau ez da haientzat hain erabilgarria izango.

Telemedikuntza oso erabilgarria eta erosoa izan daiteke kasu askotan, hala nola larrialdi egoeretan, gaixotasun arinetan eta baita denbora aurrezteko ere. Hala ere, pazienteek, batez ere adinekoek, arreta handiagoa behar dute diagnostiko hobea edo zehatzagoa egiteko. Beraz, ezinbestekoa da medikuari edo osasun-zentroari kontsultatzea behar duzunean.

## C.3 Altua testu baten adibidea

Duela pare bat urte, pandemia globalak hezkuntzan, osasunean eta lanean aldaketa batzuk egitera eraman gintuen. Hala ere, gaur egun, nahiz eta jakin egoera kezkagarria dela, ikusi dugu salbuespen honetan abian jarritako konponketetako asko mantentzen ari direla. Horren adibide da herri-adinetarako ezarri zen telemedikuntza-sistema. Edonola ere, beharrezkoa al da oraindik kontsulekin aurrera jarraitzea ikuspegi garrantzitsu horretan?

Egia da, nire ustez, medikuarekin harremanetan jarraitzeko eragozpena paregabea dela, abantaila batzuk baititu. Hasteko, onartu behar dut bazter batean gaixorik gaudenean egin nahi dugun azken gauza ohetik irtetea dela, komunera joatea, arropa janztea eta medikuarengana joatea, oro har. Gure aurrean dagoen pazienteak medikuarekin ordu erdi bat hitz egiten ematen badu ere, gure atzean zain dauden guztiak beranduago sartuko lirateke. Lehenik eta behin, telefono mugikor bat behar da deia jasateko eta, jakina, nola funtzionatzen duen jakiteko. Kasu honetan, zaharrek arazo asko izan dituzte, ez dira ondo konpontzen teknologia berriekin. Nolanahi ere, esan beharra dago telefono mugikorrik ez badugu, aukera ematen digutela telekontsultarako deia etxean egiteko. Horrek, jakina, badu eragozpena, etxean egotea beharrezkoa baita mediku batekin hitz egin nahi izanez gero.

Gainera, aurrez aurreko kontaktua ez edukitzeak medikuak gaizki diagnostikatzea eragin dezake. Askotan, funtsezkoa da medikuak pazientea ikustea sintomak aztertzeko eta gaixotasunaren eta/edo minaren gakoa eztabaidatzeko. Bestalde, kasu askotan, gaixoek ez dakite zer den beren mina, are gutxiago nola deskribatu egiten ari direna. Beraz, telemedikuntzaren bidez kalitate txarreko zerbitzu bat ematen dela uste dut, nahiz eta aurretik pentsatu dugun horiek ez direla oso kasu larriak (hasieratik gripea zela pentsa genezakeelako eta, aitzitik, biriketako minbizia izan genuen azkenean).

Beraz, esandako guztia laburbiltzeko, esan liteke herritarrek kalitatezko zerbitzua merezi dutela, eta eskubide hori beharrezkoa da.